\documentclass[lettersize,journal]{IEEEtran}
\usepackage{amsmath,amsfonts}
\usepackage{algorithmic}
\usepackage{algorithm}
\usepackage{array}
\usepackage[caption=false,font=normalsize,labelfont=sf,textfont=sf]{subfig}
\usepackage{textcomp}
\usepackage{stfloats}
\usepackage{url}
\usepackage{verbatim}
\usepackage{graphicx}
\usepackage{cite}
\usepackage{multirow}
\usepackage{booktabs}
\usepackage[pagebackref=true,breaklinks=true,letterpaper=true,colorlinks,bookmarks=false]{hyperref}
\begin{document}

\title{Counting Varying Density Crowds Through Density Guided Adaptive Selection CNN and Transformer Estimation}

\author{Yuehai~Chen, 
        Jing Yang, 
        Badong Chen, ~\IEEEmembership{Senior Member,~IEEE}
        Shaoyi~Du, ~\IEEEmembership{Member,~IEEE}

\thanks{This work was supported by the National Key Research and Development Program of China under Grant No. 2020AAA0108100, the National Natural Science Foundation of China under Grant No. 62073257, 62141223, and the Key Research and Development Program of Shaanxi Province of China under Grant No. 2022GY-076.}

\thanks{Yuehai~Chen and Jing Yang are with School of Automation Science and Engineering, Faculty of Electronic and Information Engineering, Xi’an Jiaotong, Xi’an 710049, China (e-mail: cyh0518@stu.xjtu.edu.cn, jasmine1976@xjtu.edu.cn).
}

\thanks{Badong Chen and Shaoyi~Du is with Institute of Articial Intelligence and Robotics, College of Articial Intelligence, Xi’an Jiaotong University, Xi’an, Shanxi 710049, China (e-mail: chenbd@xjtu.edu.cn, dushaoyi@gmail.com)}

\thanks{Corresponding author: Shaoyi Du. Yuehai Chen and Jing Yang contributed equally to this work.}
}

\markboth{Journal of \LaTeX\ Class Files,~Vol.~14, No.~8, August~2021}%
{Shell \MakeLowercase{\textit{et al.}}: A Sample Article Using IEEEtran.cls for IEEE Journals}


\maketitle
\begin{abstract}
In real-world crowd counting applications, the crowd densities in an image vary greatly. When facing density variation, humans tend to locate and count the targets in low-density regions, and reason the number in high-density regions. We observe that CNN focus on the local information correlation using a fixed-size convolution kernel and the Transformer could effectively extract the semantic crowd information by using the global self-attention mechanism. Thus, CNN could locate and estimate crowds accurately in low-density regions, while it is hard to properly perceive the densities in high-density regions. On the contrary, Transformer has a high reliability in high-density regions, but fails to locate the targets in sparse regions.

Neither CNN nor Transformer can well deal with this kind of density variation. To address this problem, we propose a CNN and Transformer Adaptive Selection Network (CTASNet) which can adaptively select the appropriate counting branch for different density regions. Firstly, CTASNet generates the prediction results of CNN and Transformer. Then, considering that CNN/Transformer is appropriate for low/high-density regions, a density guided adaptive selection module is designed to automatically combine the predictions of CNN and Transformer. Moreover, to reduce the influences of annotation noise, we introduce a Correntropy based optimal transport loss. Extensive experiments on four challenging crowd counting datasets have validated the proposed method.
\end{abstract}


\section{Introduction}
\begin{figure}[t]
\begin{center}
\includegraphics[width=0.9\linewidth]{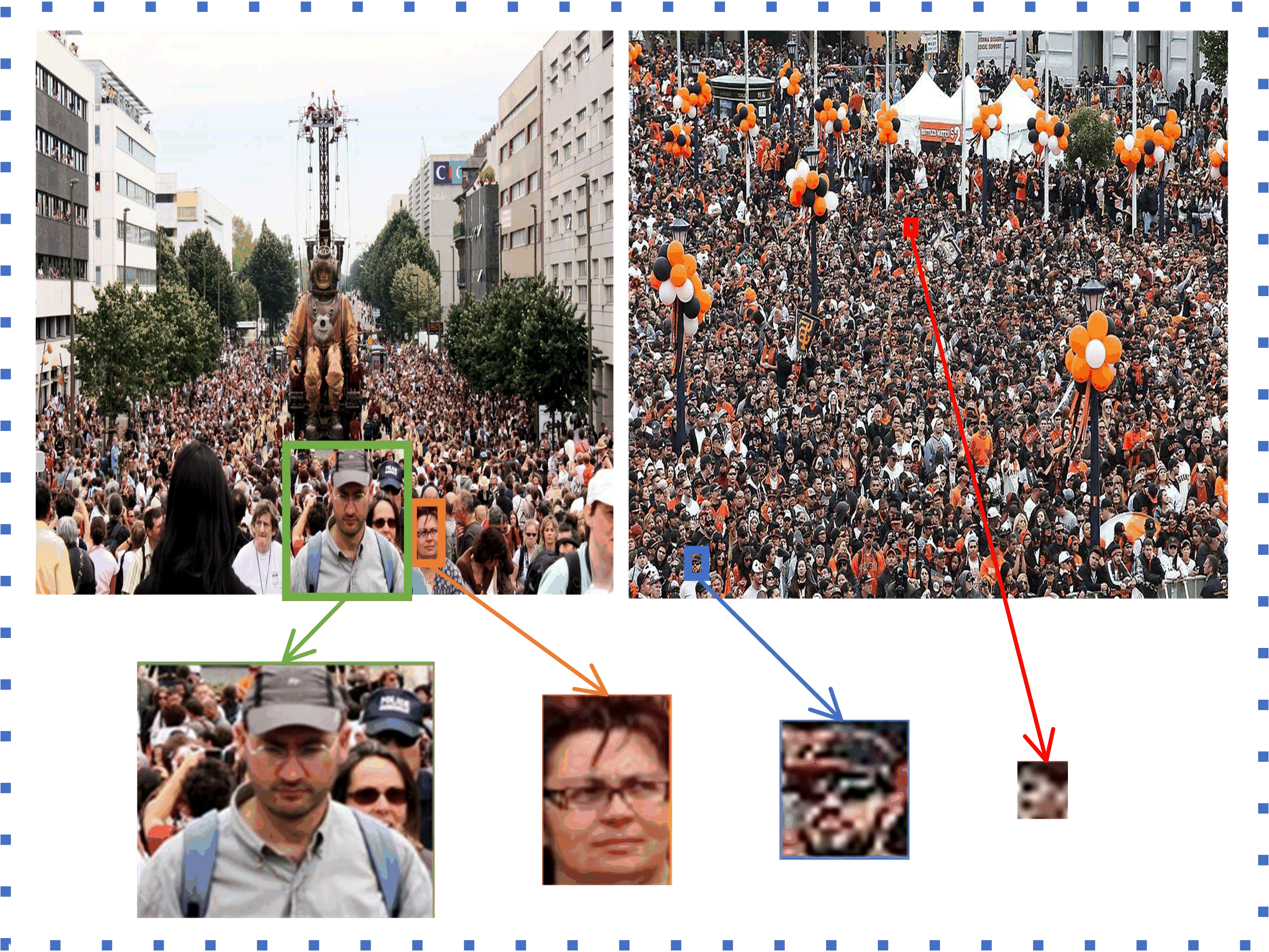}
\end{center}
   \caption{There are large scale variations in the same scene and different scenes. The intra-scene and inter-scene variations in appearance, scale, and perspective make the problem extremely difficult.}
\label{fig:introduce1}
\end{figure}

\IEEEPARstart{D}{ense} crowd counting is an important topic in computer vision. Especially after the outbreak of coronavirus disease (COVID-19), it plays a more essential role in video surveillance, public safety, and crowd analysis\cite{TCSVT1, TCSVT2, TCSVT3, TCSVT4, TNNLS2, TNNLS3, TNNLS4, TNNLS5}. Most recent state-of-the-art works generate a pseudo density map by smoothing the sparse annotated points and then train a CNN model by regressing the value at each pixel in this density map. However, as shown in Figure \ref{fig:introduce1}, a major challenge for the task is the extremely large scale variation of crowds, which arises from the wide viewing angle of cameras and the 2D perspective projection.

In recent years, numerous methods for handling large scale variations have been proposed. One feasible workaround is to adopt a multi-column convolution neural network (CNN), which aggregates several branches with different receptive fields for extracting multi-scale features \cite{MCNN, switch-CNN}. The scale variations in images are continuous, while these methods only consider several discrete receptive fields. They are unable to cover the continuous scale variation, which leads to feature redundancy among different branches \cite{CSRNet}. Considering simplifying network architecture, some methods deploy single and deeper CNNs to combine features from different layers \cite{SaCNN}. They usually aggregate the features from different layers in a scale-agnostic way, which may lead to an inconsistent mapping between feature levels and target scales \cite{SASNet}. The two kinds of aforementioned methods benefit from the multi-scale feature representations and have achieved inspiring performances. However, due to the limited respective field and presentation ability of CNN, it is still very difficult to accurately estimate the count in sparse and dense regions at the same time.

In the real world, humans would adopt appropriate counting modes for different density regions: they would accurately locate and count the targets in sparse regions and reason the target number in dense regions \cite{DecideNet}. Motivated by the human  counting behaviors, an ideal counting method should have an adaptive ability to choose the appropriate counting mode according to the crowd density. To be specific, it is expected to count through localizing targets in low density regions; whereas, in congested regions, it should behave in an inference mode. 

\begin{figure}[t]
\begin{center}
\includegraphics[width=1.0\linewidth]{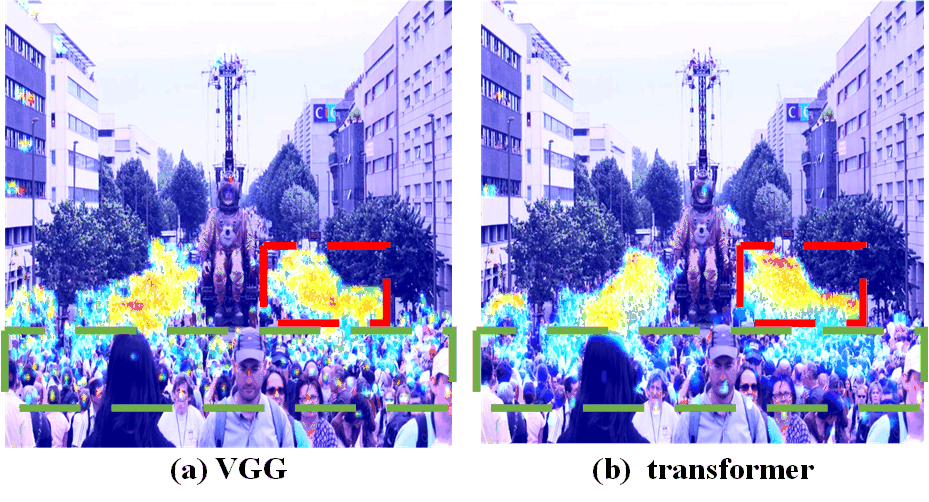}
\end{center}
   \caption{Visualization of the predicted results on a image from (a) VGG network and (b) Transformer network.}
\label{fig:introduce2}
\end{figure}

CNN constructs a powerful broad filter by focusing on the local information correlation, and show effectiveness in computer vision task. But it is usually too dependent on certain local information, which may lead to the unreliability of the extracted semantic information. Unlike the CNN, the Transformer adopts the Multi-Head Self-Attention mechanism which could provide the global receptive field. Such a Multi-Head Self-Attention mechanism can understand the relationship of different regional semantics in the entire image. In Figure \ref{fig:introduce2}, we visualize the predicted results of CNN and Transformer networks to show their differences. As shown in the green dash framework of Figure \ref{fig:introduce2} (a), the CNN estimation could accurately locate and count targets in sparse crowds, which is reliable. On the other hand, in crowded regions in the red dash framework, the corresponding object sizes tend to be very small and hard to locate. The CNN estimation is unable to accurately reflect the density in these regions, leading to a wrong estimation. While the Transformer network could perform better on the occasion, since the Transformer estimation has significantly different response intensities in different density regions. This means that the Transformer can effectively perceive the density of the area. However, in the sparse-crowd regions in the green dash framework of Figure \ref{fig:introduce2} (b), the response position of Transformer estimation is inconsistent with the target position.

Based on the above analysis, we find that the CNN and Transformer networks have their different strengths on different crowd densities regions. The CNN based method could localize and count each person precisely in the low density regions since they concentrate on the local pixel correlation. However, its reliability degenerates in crowded regions due to the inaccuracy of regional density information perception. The Transformer based approach is preferred for congested scenes. Without localization information for each target, applying them to sparse scenes may lead to wrong estimation. Motivated by this understanding, we propose a novel crowd counting framework named CNN and Transformer Adaptive Selection Network (CTASNet), which is capable of adaptively locating the target in the low density regions and perceiving the crowd density in the high density regions.

To be specific, for a given image, the CTASNet first generates two kinds of crowd density maps through CNN and Transformer networks, respectively. To adaptively decide the different counting modes for the sparse and dense regions, a density guided Adaptive Selection Module (ASM) is proposed to obtain the final prediction through automatically selecting CNN/Transformer estimations in low/high density regions, respectively. Note that point annotation is widely adopted in crowd datasets, which is sparse and could only occupy a pixel of the entire human head. There are unavoidable annotation errors. To alleviate this issue, we design a transport cost function based on correntropy \cite{Correntropy-1} in an optimal transport framework that could explicitly tolerate the annotation errors.  

In summary, we make the following contributions:
\begin{itemize}

\item To model the different counting modes of humans in sparse and dense regions, we design a CNN and Transformer Adaptive Selection Network for crowd counting.

\item We propose a density guided Adaptive Selection Module to automatically choose CNN/Transformer  network estimations for low/high density regions.

\item We design a transport cost function based on correntropy in an optimal transport framework to explicitly tolerate the annotation errors. 

\item We conduct extensive experiments on four datasets to demonstrate the superiority of our method against state-of-the-art competitors.
\end{itemize}

\section{Related works}

\subsection{Crowd Counting}
Traditional crowd counting algorithms are mainly divided into two categories: detection-based methods \cite{traditional-detetcion1, traditional-detetcion2} and regression-based methods \cite{traditional-regression1, traditional-regression2, traditional-regression3}. These methods are mostly based on hand-crafted features which are specially designed by domain experts. When faced with complex scenes such as congestion and scale variation, the performances of these hand designed methods are disappointing.

\textbf{Scale Variation.} To achieve accurate crowd counting in complex scenarios, recent attention has shifted to deep learning. The common way to cope with large scale variation is to obtain a richer feature representation \cite{SASNet}. MCNN designed a Multi-Column Neural Network to estimate crowd numbers accurately from different perspectives with three branches \cite{MCNN}. Based on MCNN, Switching CNN trains a switch classifier to select the best branch for density estimation \cite{switch-CNN}. DADNet uses different dilated rates in each parallel column to obtain multi-scale features \cite{DADNet}. \cite{wang2022crowd} used Inception-v3 as the backbone and proposed a novel curriculum loss function to resolve the scale variance issue.
\cite{MSFANet} aggregated different modules to adaptively encodes the multi-scale contextual information for accurate counting in both dense and sparse crowd scenes. In summary, these methods employ multiple branches architecture to address the scale variation problem, which may introduce significant feature redundancy \cite{CSRNet}.

\textbf{Context.} In contrast to those methods designing specific architectures, recent methods concentrate on incorporating related information like high-level semantic information and contextual information into networks for obtaining rich feature representations. CrowdNet used a combination of high-level semantic information and low-level features for accurate crowd estimation \cite{ADCrowdNet}. \cite{Cnn-based} incorporated a high-level prior into the density estimation network enabling the network to learn globally relevant discriminative features for lower count error. \cite{shang2016end, RANet} makes use of contextual information to predict both local and global counts. This kind of method has been demonstrated effective but is still difficult to accurately estimate the count in the sparse and dense regions at the same time. The reason may be that the fixed-size convolution kernel of CNN would introduce a limited receptive field.

\textbf{Loss function.} An appropriate loss measure can help to improve the counting performance. BL \cite{BL} proposes a point-wise Bayesian loss function between the ground-truth point annotations and the aggregated dot prediction generated from the predicted density map. DM-Count \cite{DM-Count} design a balanced Optimal Transport loss with an L2 cost function, to match the shape of the two distributions. Based on DM-Count \cite{DM-Count}, GL and UOT \cite{GL, UOT} develop an unbalanced Optimal Transport loss for crowd counting, which could output sharper density maps. Overly strict pixel-level spatial invariance would cause overfit noise in the density map generation. \cite{cheng2022rethinking} use locally connected Gaussian kernels to replace the original convolution filter to overcome the annotation noise in the feature extraction process.

\textbf{Crowd Localization.} To perform counting, density map estimation, and localization simultaneously, \cite{wang2021self} uses point-level annotations to train a typical object detector. \cite{gao2020learning} proposed a straightforward crowd localization framework that obtains the positions and the crowd counts by segmenting crowds into independent connected components. \cite{song2021rethinking} propose a purely point-based framework for joint crowd counting and individual localization. \cite{wang2021dense} consider the counting and localization tasks as a pixel-wise dense prediction problem and integrate them into an end-to-end framework. These crowd localization methods have achieved inspiring performances on the crowd counting task. However, they may have limited performance on the extremely dense datasets (e.g., UCF-QNRF \cite{UCF-QNRF} and UCF\_CC\_50 \cite{UCF-CC-50} dataset). Different from these localization methods, we design an adaptive selection module making CNN and Transformer models appropriate for estimating less dense and more dense regions, respectively, which could achieve better performance on extremely dense datasets.

\subsection{Transformer}

Transformers were introduced by \cite{NIPS2017_3f5ee243} as a new attention-based building block for machine translation. Dynamic attention mechanism and global modeling ability enable Transformer to exhibit strong feature learning ability \cite{transcrowd}. In recent years, Transformer has become comparable to CNN methods in computer vision \cite{Swin_Transformer}. Specifically, DETR firstly utilizes a CNN backbone to extract the visual features, followed by the Transformer for the object detection \cite{DETR}. ViT is the first one that applies Transformer-encoder to images patch and demonstrates outstanding performances \cite{ViT}. SETR extended a pure Transformer from image classification to a spatial location task of semantic segmentation \cite{SETR}. More recently, the self attention mechanism and global modeling ability have boosted the effective applications of Transformer in various tasks such as object tracking and video classification \cite{Transformer_Object_Tracking1, Transformer_Object_Tracking1}.

However, since the Transformer contains no recurrence and no convolution, the self-attention mechanism in the Transformer does not explicitly model relative or absolute position information \cite{position_encodings}. A sub-optimal approach is to add "positional encodings" to the input embeddings at the bottoms of the encoder and decoder stacks \cite{NIPS2017_3f5ee243}. Convolution can implicitly encode absolute positions, especially zero padding, and borders act as anchors to derive spatial information \cite{cnn_position1, cnn_position2}.

\subsection{Correntropy}

Correntropy, a novel similarity measure, is defined as the expectation of a kernel function between two random variables. It has been successfully applied in robust machine learning and signal processing to combat large outliers \cite{Correntropy-1}. For two different random variables $X$ and $Y$, the correntropy between $X$ and $Y$ is defined as:

\begin{equation}
V(X, Y)=\mathbf{E}[\kappa(X, Y)]=\int \kappa(x, y) d F_{\mathrm{XY}}(x, y)
\end{equation}

\noindent where $\mathbf{E}$ is the expectation operator, $\kappa(x, y)$ is a shift-invariant Mercer kernel, and $F_{X Y}(x, y)$ denotes the joint distribution function of $(X, Y)$. The most popular kernel used in correntropy is the Gaussian kernel:

\begin{equation}
\kappa_{\sigma}(x, y)=\frac{1}{\sqrt{2 \pi} \sigma} \exp \left(-\|x-y\|^{2} / 2 \sigma^{2}\right)
\end{equation}

\noindent where the $\sigma>0$ denotes the kernel size (or kernel bandwidth).

\section{Proposed Method}

\begin{figure*}
\begin{center}
\includegraphics[width=0.95\linewidth]{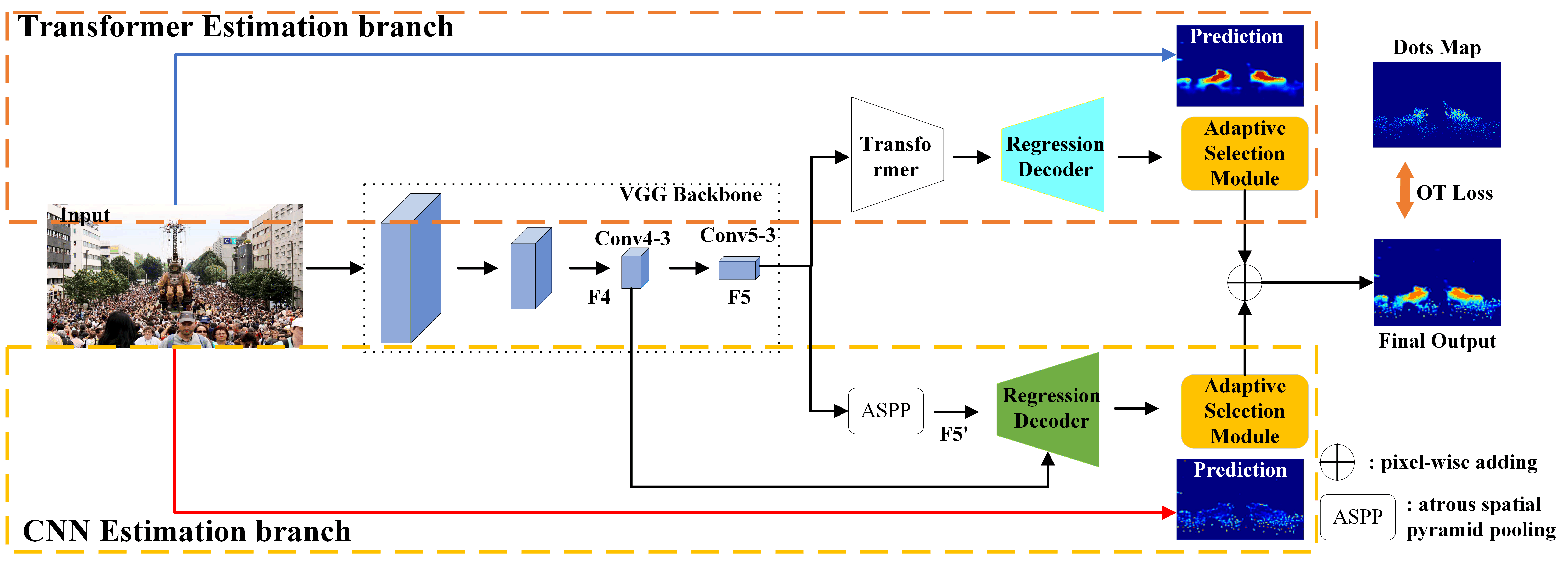}
\end{center}
   \caption{The framework of CNN and Transformer Adaptive Selection Network.}
\label{fig:framework}
\end{figure*}

In this section, we first describe the proposed crowd counting framework, CNN and Transformer Adaptive Selection Network which is shown in Figure \ref{fig:framework}. Then we will present a novel density guided Adaptive Selection Module (ASM), which could automatically choose the estimations of CNN and Transformer for different density regions. Finally, we propose a transport cost function based on correntropy in an optimal transport framework, to explicitly tolerate the annotation errors. 

\subsection{Overview}

Figure \ref{fig:framework} presents an overview of the framework. There are two parallel branches in the proposed framework: The transformer Estimation branch and the CNN Estimation branch. For each image $I$, we first use the first 13 convolution layers in the VGG16 backbone to extract the high-level feature presentations {F4, F5}. In the Transformer estimation branch, the deep feature F5 is flattened and transmitted into a Transformer encoder. Then, a regression decoder is utilized to predict the final density map $D_{t}$. In the CNN estimation branch, the top feature F5 is passed through an Atrous spatial pyramid pooling (ASPP) \cite{ASPP} module to obtain feature F5', which obtains a larger receptive field. Afterwards, a regression decoder uses concatenate and bilinear upsampling to fuse the multi-scale features F4 and F5' into the final density map $D_{c}$. Then the final output is obtained by automatically combining the predictions from CNN and Transformer branch with a designed density guided Adaptive Selection Module. Finally, we design a correntropy based OT loss to supervise the final output.

\subsection{CNN and Transformer Adaptive Selection Network}
\cite{DecideNet} shows that: in the low crowd density regions, humans would first locate the head and then count the number; while facing the high crowd density regions, they may reason about the number of people by estimating the degree of density.
To model the human different counting modes in sparse and dense regions, we design a CNN and Transformer Adaptive Selection Network for crowd counting which is presented in Figure \ref{fig:framework}. The Transformer estimation branch is responsible for the dense crowd regions, while the CNN estimation branch concentrates on predicting the sparse crowd regions. Note that Transformers require high computational costs. The cause of this challenge is the need to perform attention operations that have quadratic time and space complexity according to the context size \cite{Transformer_complex}. To reduce computational consumption, we employ a VGG 16 backbone to obtain lower-resolution feature representations which would be fed into the Transformer estimation branch and CNN estimation branch, respectively. Specifically, given the initial image $I \in \mathbb{R}^{3 \times W \times H}$, where $W$ and $H$ are the image width and height, a VGG-16 backbone generates the lower-resolution feature representations $F4 \in \mathbb{R}^{C \times \frac{W}{8} \times \frac{H}{8}}$, $F5 \in \mathbb{R}^{C \times \frac{W}{16} \times \frac{H}{16}}$.

\begin{figure}[htbp]
\begin{center}
\includegraphics[width=0.9\linewidth]{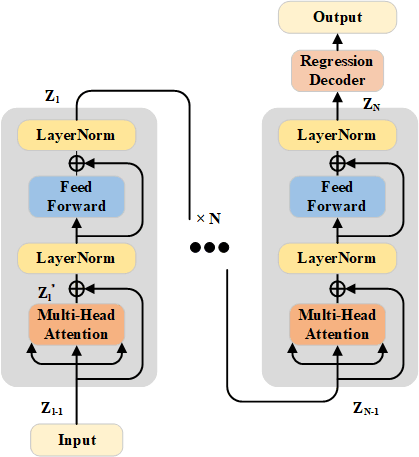}
\end{center}
   \caption{The proposed Transformer encoder consists of Multi-Head Attention, LayerNorm (LN) and a feed forward network (FFN).}
\label{fig:Transformer_framework}
\end{figure}

\hspace*{\fill} \\ 
\noindent \textbf{Transformer Estimation Branch: }
The Transformer encoder adopts self-attention layers, which can connect all pairs of input positions to consider the global relations of current features. As a result, the Transformer encoder could understand the relationship of different regional semantics in the entire image, which is beneficial to perceive the regional density. It is suitable to use a Transformer encoder to model human reasoning in dense regions.

To be specific, we only adopt the Transformer encoder with a regression decoder. As shown in Figure \ref{fig:Transformer_framework}, the proposed Transformer encoder consists of Multi-Head Attention (MHA), Layer Normalization (LN), and a feed forward network (FFN). Meanwhile, the residual connection is also employed. The output of the Transformer encoder is calculated by:

\begin{equation}
Z_{l}^{\prime}=MHA\left(Z_{l-1}\right)+Z_{l-1}
\end{equation}

\begin{equation}
Z_{l}=LN\left( FFN\left(L N\left(Z_{l}^{\prime}\right)\right)+L N\left(Z_{l}^{\prime}\right)\right)
\end{equation}

\noindent where $Z_{l}$ is the output of the $l_{th} \in {1,..., N}$ Transformer encoder, $N$ is the Transformer encoder number. $Z_{0}$ is the input of the first Transformer layer. To be specific, the Transformer encoder expects a sequence as input, hence we flatten the spatial dimensions of $F5 \in \mathbb{R}^{C \times \frac{W}{16} \times \frac{H}{16}}$ into one dimension, resulting in a $C \times \frac{WH}{256}$ feature $Z_{0}$. 

The MHA is the self-attention layer of the Transformer, which enables to measure the similarity of all input pairs to build the global relations of current features. At the MSA, the input consists of query (Q), key (K), and value (V), which are computed from $Z_{l-1}$: 

\begin{equation}
Q= Z_{l-1} W_{Q}, \quad K=Z_{l-1} W_{K}, \quad V=Z_{l-1} W_{V}
\end{equation}

\begin{equation}
MHA\left(Z_{l}\right)=\operatorname{softmax}\left(\frac{Q K^{T}}{\sqrt{d}}\right) V
\end{equation}

\noindent where $\frac{1}{\sqrt{d}}$ is a scaling factor based on the vector dimension $d$. $W^{Q}, W^{K}, W^{V} \in \mathbb{R}^{d \times d}$  three learnable weight matrices for projections.

The FFN contains two linear layers with a RELU activation function. Specifically, the first linear layer of FFN expands the feature embedding dimension from $d$ to $4d$, while the second layer shrinks the dimension from $4d$ to $d$. Finally, the output $Z_{N}$ of the Transformer encoder is fed into a regression decoder to generate the estimated density map $D_{t}$.

Since the Transformer encoder regards the input as a disordered sequence and indiscriminately considers all correlations among features, the obtained feature is position-agnostic. As a result, the prediction of the Transformer cannot have the ability to locate targets in sparse regions.

\begin{figure*}
\begin{center}
\includegraphics[width=0.95\linewidth]{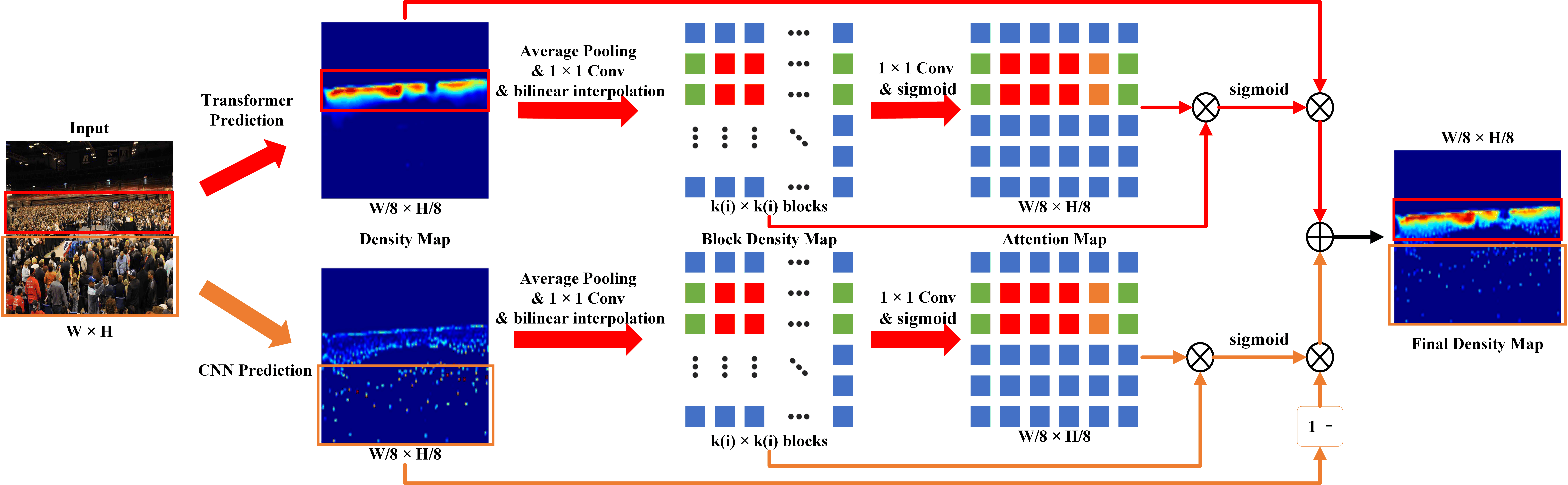}
\end{center}
   \caption{The detail structure of the proposed Adaptive Selection Module.}
\label{fig:framework1}
\end{figure*}

\hspace*{\fill} \\ 
\noindent \textbf{CNN Estimation Branch: }
Convolution is found to implicitly encode absolute positions, using zero padding, and borders act as anchors \cite{cnn_position2}. Therefore, we proposed a CNN Estimation Branch to focus on the sparse crowd regions. As illustrated in Figure \ref{fig:framework1}, the heads in sparse regions are large, thus we firstly input the feature presentation F5 into an Atrous spatial pyramid pooling (ASPP) \cite{ASPP} module to obtain the scale-aware contextual feature F5'. The ASPP module applies dilated convolution with different rates (1,6,12,18) to the obtained feature F5 for large receptive fields. To obtain accurate location information in the regression decoder, we upsample the scale-aware contextual feature F5' and concatenate it with the lower feature F4. Finally, the concatenated features are used to generate the final density map $D_{c}$.

\subsection{Density Guided Adaptive Selection Module}
After the predicted density maps {$D_{t}$, $D_{c}$} are acquired, we try to select the proper prediction for different density regions. The most common selection approach is to minimize counting error for a specific patch. However, since we could not obtain the ground truth during inference in advance, it is infeasible to decide which prediction to select for the patch. As discussed above, Transformer Estimation Branch is suitable to reason the number in high-density regions and CNN Estimation Branch could locate and count the target in the low-density regions. Therefore, we proposed a density guided adaptive selection module, which can automatically choose proper prediction manners for different density regions.

In detail, as illustrated at the Figure \ref{fig:framework1}, we compute the predicted density map $D_{i}, i\in c, t$ as:

\begin{equation}
\mathbf{d}_{bi}=\mathcal{F}_{i1}\left(P_{a v e}\left(D_{i},k(i)\right), \theta_{i1}\right)
\end{equation}

\begin{equation}
\mathbf{d}_{ati}=\mathcal{F}_{i2}\left(\mathbf{d}_{bi}, \theta_{i2}\right)
\end{equation}

\noindent where $\mathbf{d}_{bi}$ is the generated upsampled block density map and $\mathbf{d}_{ati}$ is the generated normalized attention map. The function $F_{i1}$ contains a 1 × 1 convolution network and bilinear interpolation. The function $F_{i2}$ consists of a 1 × 1 convolution network and a sigmoid function. $\theta_{ij}$(j ={1, 2}) is the learned parameter in the 1 × 1 convolution of function $F_{ij}$(j ={1, 2}). $P_{a v e}\left(.,k(i)\right)$ averages the predicted density maps into $k(i) \times k(i)$ blocks. We do this because the scale in a certain region is similar and the value in the block density map represents the regional density. We apply the Transformer/CNN prediction results in the high/low density regions.
In practice, we use $k(i) = 6$ for averaging pooling since it shows better performance compared with other settings ($k(i)=1, 2, 4, 8$). To make the block density map $P_{a v e}\left(D_{i},k(i)\right)$ obtain the same size of the predicted density map, we adopt the function $\mathcal{F}_{i1}$. To obtain a normalized attention map $\mathbf{d}_{ati}$, the function $\mathcal{F}_{i2}$ is employed. With such an operation, the obtained attention map could highlight the high-density region.

To further increase the difference between dense and sparse regions, we multiply the upsampled block density map $\mathbf{d}_{bi}$ and attention map $\mathbf{d}_{ati}$, and use sigmoid normalization to obtain the final attention map $\mathbf{A}_{ati}$. 

\begin{equation}
\mathbf{A}_{ati} = sigmoid \left(  \mathbf{d}_{bi} \times \mathbf{d}_{ati} \right)
\end{equation}

The value in the final attention map $\mathbf{A}_{ati}$ reflects the regional density. Our goal is to use the Transformer estimation branch in the dense regions and adopt the CNN estimation branch in sparse regions. Therefore, the final density map $\hat{\mathbf{z}}$ is computed by:

\begin{equation}
\hat{\mathbf{z}} = \mathbf{A}_{att} \times D_{t}+ \left(1 -\mathbf{A}_{atc} \right) \times D_{c}
\end{equation}

As a result, this would make the network automatically choose the proper prediction manner for different density regions.

\subsection{The Optimal Transport Loss based on Correntropy}
For achieving target localization in sparse regions and optimizing the entire network, we adopt the loss function proposed in DM-Count \cite{DM-Count}:

\begin{equation}
\ell(\mathbf{z}, \hat{\mathbf{z}})=\ell_{C}(\mathbf{z}, \hat{\mathbf{z}})+0.1 \ell_{O T}(\mathbf{z}, \hat{\mathbf{z}})+0.01 \|\mathbf{z}\|_{1} \ell_{T V}(\mathbf{z}, \hat{\mathbf{z}})
\end{equation}

\noindent where $\|\mathbf{z}\|$ denote the vectorized dot-annotation map, $\|\hat{\mathbf{z}}\|$ is the vectorized predicted density map. $\ell_{C}(\mathbf{z}, \hat{\mathbf{z}})=\left|\|\mathbf{z}\|_{1}-\|\hat{\mathbf{z}}\|_{1}\right|$ is the counting loss, $\|\cdot\|_{1}$ denote the $L_{1}$ norm of a vector.

\begin{equation}
\ell_{OT}(\mathbf{z}, \hat{\mathbf{z}}) \stackrel{\text { def. }}{=} \min _{\mathbf{T} \in \mathbf{U}(\mathbf{z}, \hat{\mathbf{z}})}\langle\mathbf{C}, \mathbf{T}\rangle \stackrel{\text { def. }}{=} \sum_{i, j} \mathbf{C}_{i, j} \mathbf{T}_{i, j}
\end{equation}

\noindent where $\ell_{OT}$ is the Optimal Transport (OT) loss. The ground truth $\mathbf{z}$ is the dot map. $\mathbf{C} \in \mathbb{R}_{+}^{n \times m}$ is the transport cost matrix, whose item $C_{i j}=c(\mathbf{z}, \hat{\mathbf{z}})$ measures the cost for moving probability mass on pixel $\mathbf{z}$ to pixel $\hat{\mathbf{z}}$. $\left\{\mathbf{T} \in \mathbb{R}_{+}^{n \times m}: \mathbf{T} \mathbf{1_{n}}={\mathbf{z}}, \mathbf{T}^{T} \mathbf{1_{m}}=\hat{\mathbf{z}}\right\}$ ($\mathbf{1_{n}}$ is the transport matrix, which assigns probability masses at each location $\mathbf{z}$ to $\hat{\mathbf{z}}$ for measuring the cost. $\mathbf{U}$ is the set of all possible ways to transport probability masses from $\mathbf{z}$ to $\hat{\mathbf{z}}$.

\begin{equation}
\ell_{T V}(\mathbf{z}, \hat{\mathbf{z}})=\left\|\frac{\mathbf{z}}{\|\mathbf{z}\|_{1}}-\frac{\hat{\mathbf{z}}}{\|\hat{\mathbf{z}}\|_{1}}\right\|_{T V}=\frac{1}{2}\left\|\frac{\mathbf{z}}{\|\mathbf{z}\|_{1}}-\frac{\hat{\mathbf{z}}}{\|\hat{\mathbf{z}}\|_{1}}\right\|_{1}
\end{equation}
The TV loss can increase the stability of the training procedure.

\begin{figure}[htbp]
\begin{center}
\includegraphics[width=1.0\linewidth]{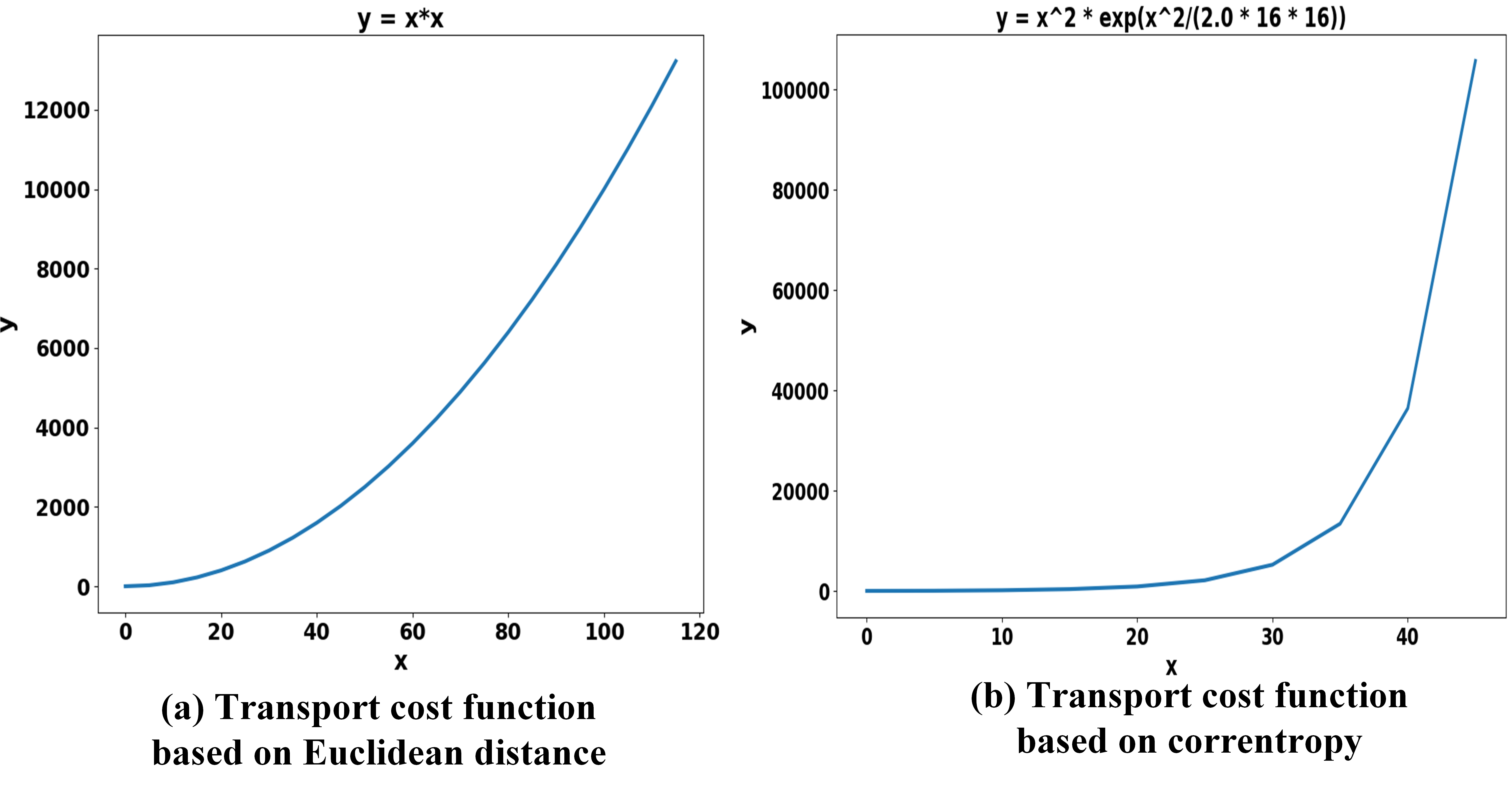}
\end{center}
   \caption{Comparison of transport cost functions based on Euclidean distance and Correntropy.}
\label{fig:kernel_function}
\end{figure}

The annotation label provided in popular datasets is in a form of sparse point annotation, which occupies a very small portion of target \cite{MAN}. The annotation noise arises from human annotation, which in general exists in this kind of annotation label. The transport cost function adopted in \cite{DM-Count} is the Euclidean distance between two pixels, $C_{i j}=L_{i j}^{2}=\left\|\boldsymbol{x}_{i}-\boldsymbol{y}_{j}\right\|_{2}^{2}$. As shown in Figure \ref{fig:kernel_function} (a), the Euclidean distance is unable to tolerate the annotation noise (i.e., the displacement of the annotated locations). If the annotated locations shifted, the distance would increase which result in a large transport cost. This means that the cost function based on Euclidean distance is sensitive to annotation noise. To alleviate negative influences by annotation noises, we propose a transport cost function based on correntropy for crowd counting:

\begin{equation}
\begin{aligned}
C_{i j} & =\frac{\left\|a_{i}-b_{j}\right\|^{2}}
{\kappa_{\sigma}(a_{i}, b_{j})}
& =\frac{\left\|a_{i}-b_{j}\right\|^{2}}
{\exp \left(- \|a_{i}-b_{j}\|^{2} / 2\sigma^{2} \right)}
\end{aligned}
\end{equation}

\noindent where $\kappa_{\sigma}\left(a_{i}, b_{j}\right)$ is a shift-invariant Gaussian kernel. As illustrated in Figure \ref{fig:kernel_function} (b), the proposed correntropy based transport cost function can alleviate negative influences caused by annotation error, since it is insensitive to annotation position offset within a certain range.

\section{Experiments}

In this section, we present experiments evaluating the proposed network, CTASNet. We first present detailed experimental setups including network architecture, training details, and evaluation metrics. Then, we compare the proposed methods with recent state-of-the-art approaches. Finally, we conduct ablation studies to verify the effectiveness of the proposed method.

\subsection{Experimental Setups}

\hspace*{\fill} \\ 
\noindent \textbf{Network Architecture.}
We adopt VGG16 as our backbone network that is pre-trained on ImageNet. Specifically, following \cite{MSFANet}, we remove the classification part of VGG-16 (fully-connected layers) and adopt the first 13 layers from VGG16 as the backbone. In the Transformer Estimation Branch, we refer to \cite{NIPS2017_3f5ee243} for the structure of Transformer encoder. The regression decoder in the Transformer Estimation Branch consists of a bilinear upsampling, two 3×3 convolution layers with 256 and 128 channels, and a 1×1 convolution layer to get the final output. In the CNN Estimation Branch, the regression decoder consists of four convolution layers with 256, 64, 32, and 1 channel, respectively. The kernel sizes of the first three layers are 3 × 3 and that of the last is 1 × 1.

\hspace*{\fill} \\ 
\noindent \textbf{Training Details.}
We first do the image augmentation using random crop and horizontal flipping. The random crop size is $512 \times 512$ in all datasets except ShanghaiTech A. As some images in ShanghaiTech A contain smaller resolution, the crop size for this dataset changes to 224 $\times$ 224. In all experiments, we use the Adam algorithm with a learning rate $10^{-5}$ to optimize the network parameters.

\hspace*{\fill} \\ 
\noindent \textbf{Evaluation Metrics.}
The widely used mean absolute error (MAE) and the mean squared error (MSE) are adopted to evaluate the performance. The MAE and MSE are defined as follows:

\begin{equation}
M A E=\frac{1}{N} \sum_{i=1}^{N}\left|\hat{C}_{i}-C_{i}\right|
\end{equation}

\begin{equation}
M S E=\sqrt{\frac{1}{N} \sum_{i=1}^{N}\left|\hat{C}_{i}-C_{i}\right|^{2}}
\end{equation}
where $N$ is the number of test images, $\hat{C}_{i}$ and $C_{i}$ are the estimated count and the ground truth, respectively.

\subsection{Comparisons with State-of-the-Arts}

\begin{table*}
\caption{Results on the ShanghaiTech, UCF\_CC\_50, UCF-QNRF and NWPU datasets.}
\begin{center}
\begin{tabular}{l|ll|ll|ll|ll|ll}
 \hline 
 \multicolumn{1}{l} {Method} & \multicolumn{2}{l}{ ShanghaiTech A } & \multicolumn{2}{l}{ ShanghaiTech B } & \multicolumn{2}{l}{ UCF\_CC\_50 } & \multicolumn{2}{l}{ UCF-QNRF } & \multicolumn{2}{l}{ NWPU} \\
& {MAE} & MSE & MAE & MSE & MAE & MSE & MAE & MSE & MAE & MSE \\
\hline MCNN\cite{MCNN} & $110.2$ & $173.2$ & $26.4$ & $41.3$ & $377.6$ & $509.1$ & $277.0$ & $426.0$ & $232.5$ & $714.6$ \\
Switch-CNN\cite{switch-CNN} & $90.4$ & $135.0$ & $21.6$ & $33.4$ & $318.1$ & $439.2$ & 228 & 445 & $-$ & $-$ \\
CSRNet\cite{CSRNet} & $68.2$ & $115.0$ & $10.6$ & $16.0$ & $266.1$ & $397.5$ & $120.3$ & $208.5$ & $121.3$ & $387.8$\\
SANet\cite{SANet} & $67.0$ & $104.2$ & $8.4$ & $13.6$ & $258.4$ & $334.9$ & $-$ & $-$& $-$ & $-$ \\
CAN\cite{CAN} & $62.3$ & 100 & $7.8$ & $12.2$ & $212.2$ & $243.7$ & 107 & 183 & $106.3$ & $\underline{386.5}$ \\
BL\cite{BL} & $62.8$ & $101.8$ & $7.7$ & $12.7$ & $229.3$ & $308.2$ & $88.7$ & $154.8$  & $105.4$ & $454.0$\\
SFCN\cite{SFCN} & $67.0$ & $104.5$ & $8.4$ & $13.6$  &  $258.4$ & $334.9$ & $102.0$ & $171.4$ & $105.7$ & $424.1$\\


ADSCNet\cite{ADCrowdNet} & $55.4$ & $97.7$ & $\textbf{6.4}$ & $11.3$  & $198.4$ & $267.3$ & $\textbf{71.3}$ & $\textbf{132.5}$ & $-$ & $-$ \\
CG-DRCN-CC-Res101\cite{JHU2} & $60.2$ & $94.0$ & $7.5$ & $12.1$  & $-$ & $-$ & $95.5$ & $164.3$ & $-$ & $-$ \\
SASNet\cite{SASNet} & $\textbf{53.6}$ & $\underline{88.4}$ & $\textbf{6.4}$ & $\textbf{10.0}$  & $\underline{161.4}$ & $\underline{234.5}$ & $85.2$ & $147.3$ & $-$ & $-$\\
NoiseCC\cite{NoiseCC} & $61.9$ & $99.6$ & $7.4$ & $11.3$  & $-$ & $-$ & $85.8$ & $150.6$ & $96.9$ & $534.2$\\
\hline 
DM-Count\cite{DM-Count} & $59.7$ & $95.7$ & $7.4$ & $11.8$  & $211.0$ & $291.5$ & $85.6$ & $148.3$ & $\textbf{88.4}$ & $388.4$\\
CTASNet(ours)  & $\underline{54.3}$ & $\textbf{87.8}$ & $\underline{6.5}$ & $\underline{10.7}$ & $\textbf{158.1}$ & $\textbf{221.9}$ & $\underline{80.9}$ & $\underline{139.2}$ & $\underline{94.4}$ & $\textbf{357.6}$\\
\hline 
\end{tabular}
\end{center}
\label{table:Comparisons}
\end{table*}

We compare our proposed methods with other state-of-the-art methods on several public crowd datasets, including ShanghaiTech A \cite{MCNN}, ShanghaiTech B \cite{MCNN}, UCF\_CC\_50 \cite{UCF-CC-50}, UCF-QNRF \cite{UCF-QNRF} and NWPU \cite{NWPU}. The quantitative results of counting accuracy are listed in Table \ref{table:Comparisons}.

\hspace*{\fill} \\ 
\noindent \textbf{ShanghaiTech A Dataset.}
ShanghaiTech A dataset is collected from the Internet and consists of 482 (300 for train, 182 for test) images with highly congested scenes. The images in ShanghaiTech A are highly dense with crowd counts between 33 to 3139. Our CTASNet achieves the lowest MSE and the second lowest MAE on ShanghaiTech A. Compared to DM-Count, CTASNet significantly boosts its counting accuracy on ShanghaiTech A. Specifically, the improvements are 9.04\% and 8.25\% for MAE and MSE.

\hspace*{\fill} \\ 
\noindent \textbf{ShanghaiTech B Dataset.}
The ShanghaiTech B dataset contains 716 (400 for train, 316 for test) images taken from busy streets in Shanghai. The images in ShanghaiTech B are less dense with the number of people varying from 9 to 578. It can be observed that the proposed CTASNet can deliver comparable results with the best method (SASNet). The effect of our CTASNet could not outperform SASNet. The reason may be that ShanghaiTech B dataset contains fewer people than other popular datasets. There are fewer extremely dense regions that are suitable for reasoning, thus our Transformer estimation branch is not so useful in these low-density scenes. Compared with methods with the DM-Count \cite{DM-Count} performance, the CTASNet reduces the MAE by 12.16\% and MSE by 9.32\%.

\hspace*{\fill} \\ 
\noindent \textbf{UCF\_CC\_50.}
UCF\_CC\_50 is an extremely dense crowd dataset but includes 50 images of different resolutions \cite{UCF-CC-50}. The numbers of annotations range from 94 to 4,543 with an average number of 1,280. Due to the limited training samples, 5-fold cross-validation is performed following the standard setting in \cite{UCF-CC-50}. As shown in Table~\ref{table:Comparisons}, our CTASNet surpasses all the other methods. In particular, our method could achieve the best performance with 5.77\% MAE and 8.68\% MSE improvement compared to SASNet \cite{SASNet} with the second best performance.

\hspace*{\fill} \\ 
\noindent \textbf{UCF-QNRF.}
UCF-QNRF is a challenging dataset that has a much wider range of counts than currently available crowd datasets \cite{UCF-QNRF}. Specifically, the dataset contains 1,535 (1,201 for train, 334 for test) jpeg images whose number ranges from 816 to 12,865. By combining the CNN and Transformer estimations, our method can achieve the second best performance with an MAE of 80.9 and MSE of 139.2. Note that our CTASNet makes a reduction of 6.34\% on MSE compared with DM-Count.

\hspace*{\fill} \\ 
\noindent \textbf{NWPU Dataset.}
The NWPU dataset is the largest-scale and most challenging crowd counting dataset publicly available \cite{NWPU}. The dataset consists of 5,109 (3,109 for train, 500 for val , 1,500 for test) images whose number ranges from 0 to 20,003. Note that the ground truth of the test image sets is not released and researchers could submit their results online for evaluation. As illustrated in Table~\ref{table:Comparisons}, our CTASNet achieves the lowest MSE and the second lowest MAE on NWPU.

\begin{figure*}
\begin{center}
\includegraphics[width=0.95\linewidth]{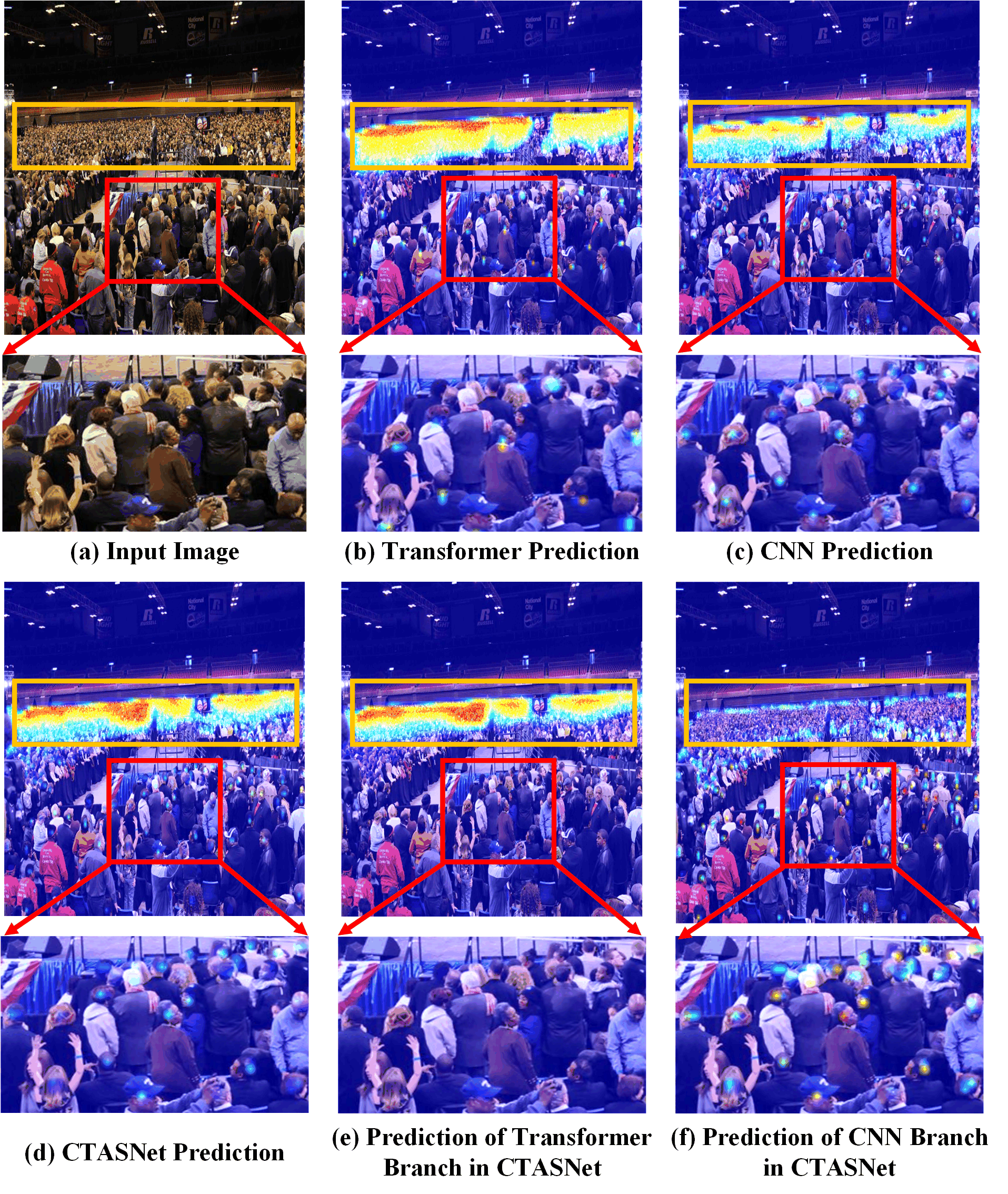}
\end{center}
   \caption{Visualization of the prediction results: (a) Input Image; (b) Prediction of Transformer; (c) Prediction of CNN; (d) Prediction of CNN and Transformer Adaptive Selection Network (CTASNet); (e) Prediction of Transformer Estimation Branch in CTASNet; (f) Prediction of CNN Estimation Branch in CTASNet.}
\label{fig:vis}
\end{figure*}

\subsection{Ablation Study}

\subsubsection{Effect of each component}
\hspace*{\fill} \\ 
\noindent \textbf{Quantitative Analysis.}
The proposed CTASNet is composed of four components: CNN Estimation Branch (CEB), Transformer Estimation Branch (TEB), Density Guided Adaptive Selection Module (ASM) and Correntropy based Optimal Transport loss (COT). We perform ablation studies on ShanghaiTech A dataset to analyze the effect of each component. We first adopt the CEB with different loss functions L2 loss \cite{CSRNet}, Bayesian Loss (BL) \cite{BL}, optimal transport (OT) loss in DM-Count \cite{DM-Count} and the proposed COT loss. Then, we use the CEB and TEB for prediction, respectively. Finally, we adopt two strategies ("concat" and ASM) to combine the predictions of CEB and TEB. To be specific, in the "concat" strategy, we concat the features generated from the encoders in CEB and TEB. Afterwards, we feed the obtained feature into the regression decoder for the final prediction. While ASM is the proposed Adaptive Selection Module in the Methods Section.
     
All detail results are illustrated in Table \ref{tab:component}. Comparing the results of CEB with different loss functions, we could observe that the CEB with our COT loss achieves the best result. Specifically, by replacing L2 loss with the proposed COT loss, the performance of CEB is improved by 16.1\% in MAE and 17.7\% in MSE. This illustrates that the proposed COT loss can significantly boost the counting accuracy.
     
One could observe that using both CNN and Transformer for the prediction can effectively improve the counting accuracy. Specifically, such a simple "concat" strategy can achieve an MAE of 56.4 and an MSE of 90.3. However, for a specific region, if we aggregate the predictions by adaptively selecting the different counting modes in different density regions, the MAE and MSE are improved to 54.3 and 87.8, denoted by the 'CEB + TEB + ASM + COT' in Table \ref{tab:component}. This significant improvement demonstrates the great potential of automatically selecting the appropriate counting mode in different density regions.
     
     \begin{table}[htbp]
\caption{Analysis of the effect of different components. All experiments are performed on ShanghaiTech A dataset.}
\begin{center}
\begin{tabular}{ccc}
\hline \text {Components} & \text { MAE } & \text { MSE } \\
\hline 
CEB + L2 & 67.6 & 113.9 \\
CEB + BL & 62.4 & 100.5 \\
CEB + OT loss in DM-Count \cite{DM-Count} & 59.2 & 96.4 \\
CEB + COT & 56.7 & 93.7 \\
TEB + COT & 61.2 & 95.4 \\
CEB + TEB + concat + COT & 56.4 & 90.3 \\
CEB + TEB + ASM + COT & 54.3 & 87.8 \\
\hline
\end{tabular}
\end{center}
\label{tab:component}
\end{table}

\hspace*{\fill} \\ 
\noindent \textbf{Qualitative Analysis.} To further demonstrate the advantage of the ASM strategy, we visualize the predictions of 'CEB+ COT', 'TEB+ COT' and 'CEB + TEB + ASM + COT' in Figure \ref{fig:vis}. Specifically, (a) is the input image; (b) is the prediction of TEB; (c) is the prediction of CEB; (d) is the prediction of CTASNet; (e) is the prediction of TEB in CTASNet; (f) is the prediction of CEB in CTASNet.

Firstly, comparing the yellow frames in (a), (b), and (c), we could observe that Transformer has significantly different response intensities in dense regions of varying degrees. This illustrates that Transformer has a strong ability to perceive the regional density. While for sparse-crowd regions in red frames, the Transformer fails to locate and count accurately the target head. CNN shows opposite characteristics in prediction: CNN could well locate and count in the sparse-crowd region but has a similar response intensity in dense-crowd regions of varying degrees. It means that CNN may be suitable for sparse-crowd region counting and Transformer may work well in dense-crowd region counting.

The (d) image in Figure \ref{fig:vis} presents our CTASNet predictions. In Figure \ref{fig:vis} (d), it is easy to observe that our CTASNet can locate targets in low-density regions and has different response intensities for different density regions. This verifies that our CTASNet could take full advantage of CNN and Transformer to achieve accurate counting on both sparse and dense regions. More specifically, in Figure \ref{fig:vis} (e) and (f), we visualize the predictions of TEB and CEB in the proposed CTASNet. One could observe that the TEB prediction only focuses on dense regions. Also, the CEB prediction mostly concentrates on sparse regions. These results fully demonstrate that our CTASNet can adaptively select the appropriate counting mode for different density regions. This is consistent with people's counting behavior.

    \begin{figure}[htbp]
    \centering
    \centerline{\includegraphics[width=1\linewidth]{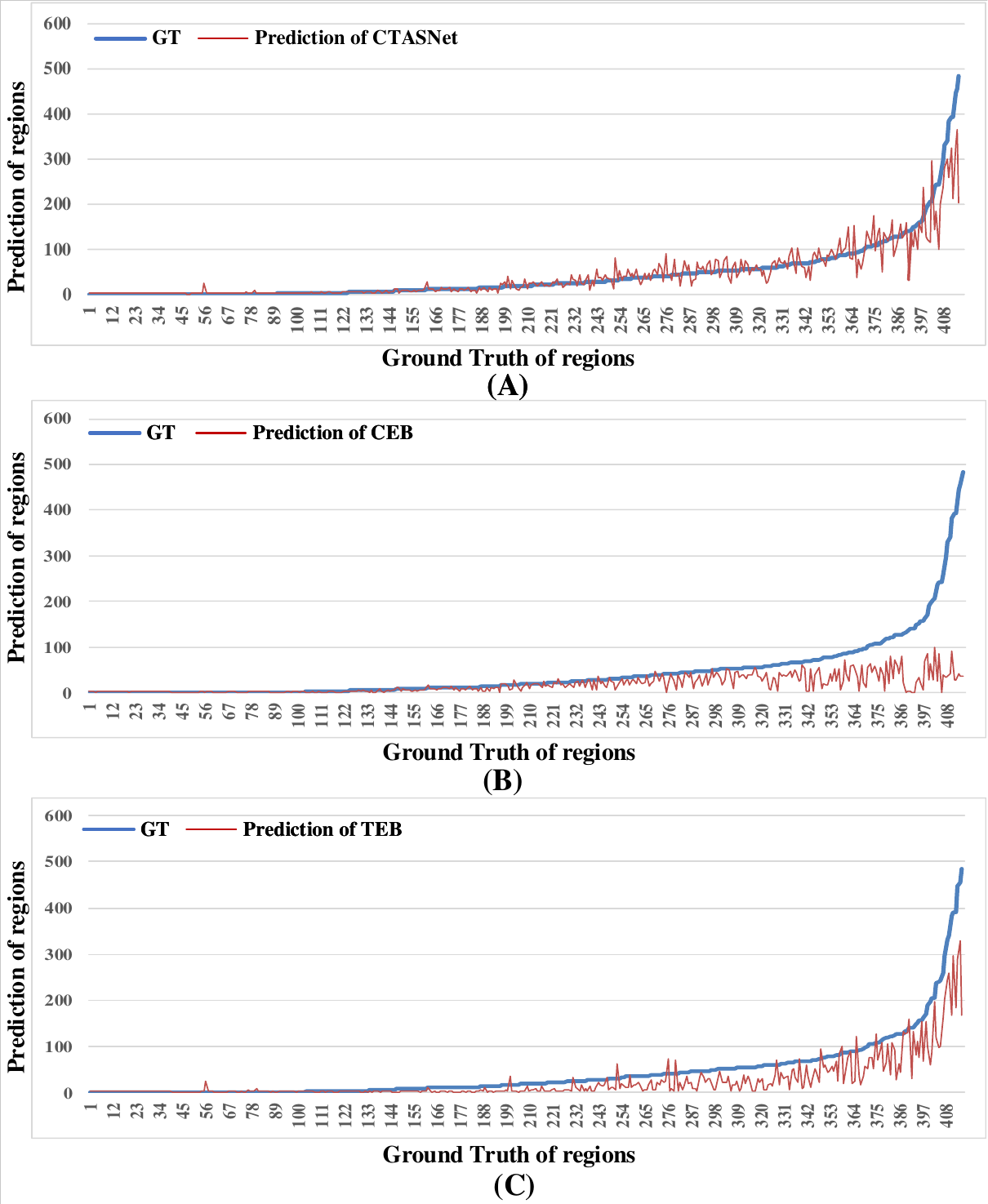}}
    \caption{Visualization of the quantitative experiment. The prediction of CTASNet is a combined prediction of CNN and Transformer. The CNN and the Transformer branches are responsible for less dense and more dense regions, respectively.}
    \label{quantitative}
    \end{figure}

\hspace*{\fill} \\ 
\noindent \textbf{Analyze the predictions of CEB and TEB.}
To further show that the CNN and the Transformer branches are responsible for less dense and more dense regions, respectively, we conduct a quantitative experiment. Specifically, we first choose a relatively dense subset (contains 26 images, about 15\% test set) from the test images (contains 181 images) in the ShanghaiTech A dataset. Then, we divide them into $4 \times 4$ blocks which contain low and high-density regions. Finally, we compare the count predictions and ground truth between CNN and Transformer branches in Figure \ref{quantitative}.
    
    As shown in Figure \ref{quantitative} (A), it could be observed that the prediction of CTASNet is close to the ground truth. While for the extremely dense regions (ground truth of region \textgreater 400), the CTASNet underestimates them. The reason may be that these extremely dense regions are too less to learn effective features for accurate estimation.
    
    Figure \ref{quantitative} (B) has presented that, when facing the less dense regions (ground truth of region \textless 250), the prediction of CEB is close to the ground truth. While for TEB in Figure \ref{quantitative} (C), we could observe that the prediction of TEB approximates 0. This means that the estimation of TEB contributes less to these sparse regions. It confirms the CNN branch is responsible for less dense regions. On the contrary, when estimating more dense regions (ground truth of region \textgreater 325), the prediction of TEB occupies a main component of the final estimation compared to the prediction of CEB. It illustrates that the Transformer branch is responsible for more dense regions.
  
  \begin{figure*}[htbp]
\begin{center}
\includegraphics[width=0.95\linewidth]{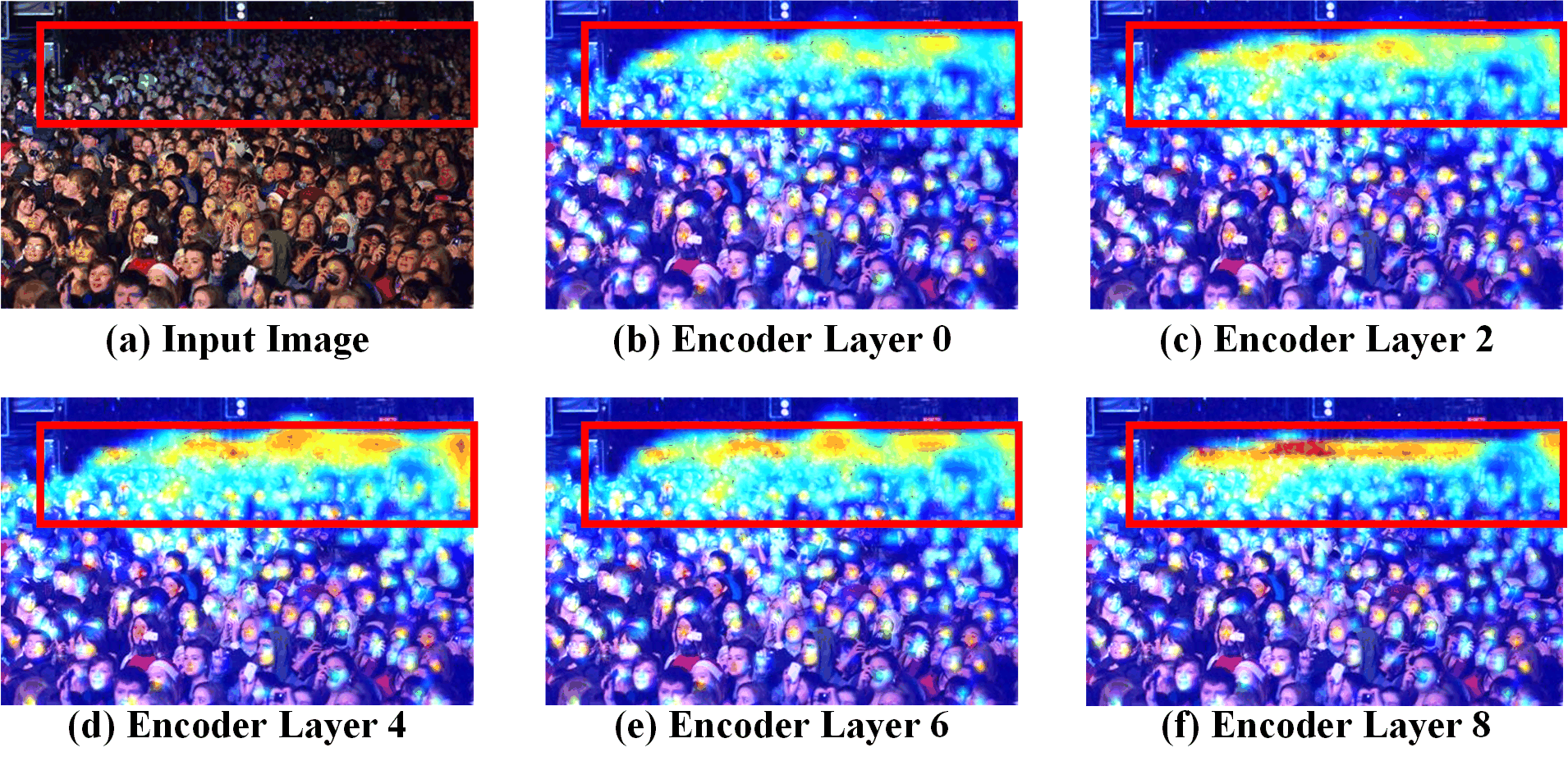}
\end{center}
   \caption{Visualization of the prediction results containing different Transformer encoder layers.}
\label{fig:transnumber}
\end{figure*}  
  
\begin{table*}[htbp]
\caption{Effect of encoder layer numbers. Each row corresponds to a model with the varied number of encoder layers. Performance gradually improves with more encoder layers.}
\begin{center}
\begin{tabular}{lccccc}
\hline layers & GFLOPS & params & Inference time & MAE & MSE \\
\hline 
0 & $86.5 \mathrm{G}$  & $19.4 \mathrm{M}$ & $2.9$ & $56.7$ & $93.7$\\
2 & $97.4 \mathrm{G}$  & $25.6 \mathrm{M}$ & $3.6$ & $56.6$ & $89.7$\\
4 & $102.2 \mathrm{G}$  & $30.3 \mathrm{M}$ & $4.0$ & $54.3$ & $87.8$ \\
6 & $107.1 \mathrm{G}$ & $35.0 \mathrm{M}$ & $4.4$ & $53.9$ & $87.2$\\
8 & $111.9 \mathrm{G}$ & $39.8 \mathrm{M}$ & $4.9$ & $52.4$ & $86.3$ \\
\hline
\end{tabular}
\end{center}
\label{tab:trans}
\end{table*}

\begin{table*}[htbp]
\caption{Performances of loss functions using different backbones on UCF-QNRF dataset. Our proposed method outperforms other loss functions.}
\begin{center}
$$
\begin{array}{lllllll}
\hline \multirow{2}*{\text { Methods }} & \multicolumn{2}{c}{\text { VGG19  }}&  \multicolumn{2}{c}{\text { CSRNet  }}& \multicolumn{2}{c}{\text { MCNN } } \\
& \text { MAE } & \text { MSE } & \text { MAE } & \text { MSE } & \text { MAE } & \text { MSE } \\
\hline \text { L2 } & 98.7 & 176.1 & 110.6 & 190.1 & 186.4 & 283.6 \\
\text { BL \cite{BL} } & 88.8 & 154.8 & 107.5 & 184.3 & 190.6 & 272.3 \\
\text { NoiseCC \cite{NoiseCC}} & 85.8 & 150.6 & 96.5 & 163.3 & 177.4 & 259.0 \\
\text { DM-Count \cite{DM-Count} } & 85.6 & 148.3 & 103.6 & 180.6 & 176.1 & 263.3 \\
\text {AL-PAPM (OURS)} & \mathbf{81.2} & \mathbf{141.9} & \mathbf{95.6} & \mathbf{162.7} & \mathbf{157.5} & \mathbf{243.3} \\
\hline
\end{array}
$$
\end{center}
\label{tab:different backbones}
\end{table*}

\subsubsection{Effect of the number of Transformer Encoder Layer}

We evaluate the importance of global feature-level self-attention by changing the number of Transformer encoder layers. Table \ref{tab:trans} reports a detail comparison. Comparing layer 0 and layer 2, we could observe that, without Transformer encoder layers, there is a significant MSE drop by 4.0 points. Moreover, the performance gradually improves with more Transformer encoder layers. Thus we conjecture that the Transformer encoder which uses global scene reasoning is useful to perceive the different regional densities in the whole image. As shown in Table \ref{tab:trans}, the layer of 8 achieves the best results. However, compared with the layer of 4, the layer of 8 brings an increase of 31.4\% in parameters and 22.5\% in inference time. By adding the layer from 4 to 8, the improvement of MSE is 1.7\% on ShanghaiTech A. Taking into account model complexity and counting accuracy, we adopt 4 Transformer Encoder layers in our proposed CTASNet.

In Figure \ref{fig:transnumber}, we visualize the prediction results of different Transformer encoder layers. By comparing the results of encoder layer 0 and other encoder layers, it could be observed that the predictions with the Transformer encoder layer have significantly different response intensities in different density regions. This verifies that the Transformer encoder using global scene reasoning is effective to perceive the different regional densities in the dense regions. 

\subsubsection{Effect of Bandwidth $\sigma$}

\begin{figure}[htbp]
\begin{center}
\includegraphics[width=1.0\linewidth]{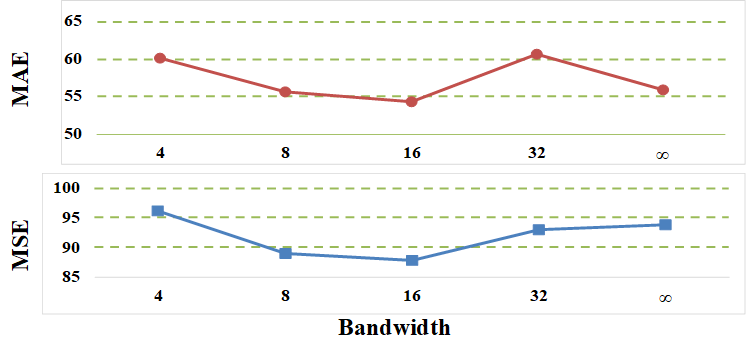}
\end{center}
   \caption{The curves of testing results for different bandwidths on ShanghaiTech A.}
\label{fig:bandwidth}
\end{figure}

We next investigate the effect of bandwidth $\sigma$ in our proposed Correntropy based OT Loss, which is designed for tolerating the annotation noise. The specific Correntropy based OT Loss is computed by:

\begin{equation}
\ell_{OT}(\mathbf{z}, \hat{\mathbf{z}}) \stackrel{\text { def. }}{=} \min _{\mathbf{T} \in \mathbf{U}(\mathbf{z}, \hat{\mathbf{z}})}\langle\mathbf{C}, \mathbf{T}\rangle \stackrel{\text { def. }}{=} \sum_{i, j} \mathbf{C}_{i, j} \mathbf{T}_{i, j}
\end{equation}

\begin{equation}
\begin{aligned}
C_{i j} & =\frac{\left\|a_{i}-b_{j}\right\|^{2}}
{\kappa_{\sigma}(a_{i}, b_{j})}
& =\frac{\left\|a_{i}-b_{j}\right\|^{2}}
{\exp \left(- \|a_{i}-b_{j}\|^{2} / 2\sigma^{2} \right)}
\end{aligned}
\label{cost}
\end{equation}

In this experiment, we tune bandwidth $\sigma$ from 4, 8, 16, 32 to $\infty$. Especially, when $p=\infty$, the Correntropy based cost function presented in Equation (\ref{cost}) converts to the classical $L_{2}$ cost function $C_{ij} = \left\|a_{i}-b_{j}\right\|^{2}$. The detail results are presented in Figure \ref{fig:bandwidth}, $\sigma = 16$ outperforms other bandwidth values. Therefore, we choose $\sigma = 16$, for all experiments on all datasets.

\subsubsection{Effect of the position embedding}

To investigate the effect of the position embedding, we conduct an ablation study on ShanghaiTech A dataset. To be specific, following \cite{NIPS2017_3f5ee243}, we use sine and cosine functions of different frequencies to achieve positional encoding. Details are presented in Table \ref{position}. Comparing the results of TEB and ‘TEB + position encoding’, we could observe that adding position encoding to TEB can improve the counting performance. While for CTASNet, embedding the positional information would reduce the counting accuracy. The reason may be that CTASNet contains convolution for final estimation, enabling the model can implicitly encode absolute positional information. Adding extra positional information would bring information redundancy. Thus, we do not introduce extra positional encoding in the Transformer of CTASNet.
    
    \begin{table}
    \begin{center}
    \caption{Effect of the position embedding. Following \cite{NIPS2017_3f5ee243}, we add sine and cosine positional encoding into Transformer Estimation Branch (TEB) and CTASNet. We adopt the proposed Correntropy based Optimal Transport loss (COT) for TEB and CTASNet.}
    \begin{tabular}{lll}
    \hline Components & \multicolumn{2}{c}{ ShanghaiTech $\mathrm{A}$} \\
    & MAE & MSE \\
    \hline TEB & $61.2$ & $95.4$ \\
    TEB + position encoding & $57.7$ & $91.7$ \\
    CTASNet & $54.3$ & $87.8$ \\
    CTASNet + position encoding & $56.6$ & $88.7$ \\
    \hline
    \end{tabular}
    \label{position}
    \end{center}
    \end{table}

\subsubsection{Comparison with different loss functions}

In Table \ref{tab:different backbones}, we compare our proposed loss function with different loss functions using different backbone networks. The pixel-wise L2 loss function measures the pixel difference between the predicted density map and the "ground-truth" density map. The Bayesian loss (BL) \cite{BL} uses a point-wise loss function between the ground-truth point annotations and the aggregated dot prediction generated from the predicted density map. The NoiseCC model \cite{NoiseCC} the annotation noise using a random variable with Gaussian distribution and derives a probability density Gaussian approximation as a loss function. DM-Count \cite{DM-Count} uses balanced OT with an L2 cost function, to match the shape of the two distributions. 

Our proposed loss function achieves the best results among all loss functions. Our loss function performs better than L2 loss since we directly adopt point annotation for supervision, instead of designing hand-craft intermediate presentation as a learning target. Compared to BL and DM-Count (both using point annotation for supervision), our loss function achieves better performances in all network architectures. Our loss function can tolerate the annotation noise. While these two loss functions ignore the noise problem existing in the annotation labeling process. Thus, our proposed loss function achieves the lowest MAE and MSE among all loss functions.

\subsection{Complexity analysis}
To evaluate the complexity of our method, we have conducted an ablation study on ShanghaiTech A dataset in Table \ref{tab6}. To exclude interference from other factors, we experimented on the same experimental environment and reported the results in ShanghaiTech A benchmark \cite{MCNN}. The parameters and FLOPs are computed with the input size of 512×512 on a single NVIDIA 3090 GPU. The inference time is the average time of 100 runs on testing a 1024×768 image sample.

As shown in Table \ref{tab6}, our model does not have an advantage in model parameters and inference speed. However, our model has achieved better performance in crowd counting. Moreover, our model could also achieve real-time crowd counting at a speed of 0.048 seconds per picture. It does not affect the application of our method in reality.

    \begin{table*}[htbp]
\caption{Comparison of the Parameters (M), FLOPs (G), Inference speed (s / 100 images) and results on the PartA dataset \cite{MCNN}. Note: The parameters and FLOPs are computed with the input size of 512×512 on a single NVIDIA 3090 GPU. The inference time is the average time of 100 runs on testing a 1024×768 sample. “FS” represents that the model is trained From Scratch.}
\begin{center}
\begin{tabular}{llllll}
\hline Method & Backbone &  Parameters(M) & FLOPs(G) & Inference speed(s) & RMSE In PartA\\
\hline 
MCNN \cite{MCNN} & FS & $\textbf{0.13}$ & $\textbf{7.05}$ & $\textbf{0.008}$ & 173.2\\
PCC-Net\cite{TCSVT3} & FS & $0.51$ & $43.87$ & $0.013$ & 124.0 \\
CSRNet\cite{CSRNet} & VGG16 & $16.26$ & $108.34$ & $0.038$ & 115.0 \\
CAN\cite{CAN} & VGG16 & $18.10$ & $114.83$ & $0.047$ & 100.0\\
SCAR\cite{SCAR} & VGG16 & $16.29$ & $108.44$ & $0.047$ & 110.0\\
SFANet\cite{MSFANet} & VGG16 & $15.92$ & $93.27$ & $0.043$& $99.3$ \\
M-SFANet\cite{MSFANet} & VGG16 & $22.88$ & $115.14$ & $0.058$& $94.5$ \\
SFCN\cite{SFCN} & ResNet101 & $38.60$ & $162.03$ & $0.096$& 104.5 \\
CTASNet(ours) & VGG16 & $30.30$ & $102.22$ & $0.040$ & $\textbf{87.8}$\\
\hline
\end{tabular}
\end{center}
\label{tab6}
\end{table*}

\section{Conclusion}
In this paper, a CNN and Transformer Adaptive Selection Network (CTASNet) has been proposed for the crowd counting problem. It is motivated by the complementary performance of CNN and Transformer based counting methods under situations with varying crowd densities. An Adaptive Selection Module is proposed to make the network automatically adopt different counting modes for different density regions. Also, to reduce the influences of annotation noise, we introduce a Correntropy based Optimal Transport loss (COT loss). We evaluate the CTASNet with proposed COT loss on four challenging crowd counting benchmarks, most of which consist of high variation in crowd densities. The experimental results confirm that our method obtains state-of-the-art performance on these datasets. 

\section*{Acknowledgments}
This work was supported by the National Key Research and Development Program of China under Grant No. 2020AAA0108100, the National Natural Science Foundation of China under Grant No. 62073257, 62141223, and the Key Research and Development Program of Shaanxi Province of China under Grant No. 2022GY-076.

\vspace{11pt}




\vspace{11pt}


\vfill


\begin{thebibliography}{100}
\providecommand{\url}[1]{#1}
\csname url@samestyle\endcsname
\providecommand{\newblock}{\relax}
\providecommand{\bibinfo}[2]{#2}
\providecommand{\BIBentrySTDinterwordspacing}{\spaceskip=0pt\relax}
\providecommand{\BIBentryALTinterwordstretchfactor}{4}
\providecommand{\BIBentryALTinterwordspacing}{\spaceskip=\fontdimen2\font plus
\BIBentryALTinterwordstretchfactor\fontdimen3\font minus
  \fontdimen4\font\relax}
\providecommand{\BIBforeignlanguage}[2]{{%
\expandafter\ifx\csname l@#1\endcsname\relax
\typeout{** WARNING: IEEEtran.bst: No hyphenation pattern has been}%
\typeout{** loaded for the language `#1'. Using the pattern for}%
\typeout{** the default language instead.}%
\else
\language=\csname l@#1\endcsname
\fi
#2}}
\providecommand{\BIBdecl}{\relax}
\BIBdecl

\bibitem{DLA}
F.~Yu, D.~Wang, E.~Shelhamer, and T.~Darrell, ``Deep layer aggregation,'' in
  \emph{Proceedings of the IEEE conference on computer vision and pattern
  recognition}, 2018, pp. 2403--2412.

\bibitem{SCAR}
J.~Gao, Q.~Wang, and Y.~Yuan, ``Scar: Spatial-/channel-wise attention
  regression networks for crowd counting,'' \emph{Neurocomputing}, vol. 363,
  pp. 1--8, 2019.

\bibitem{cheng2022rethinking}
Z.-Q. Cheng, Q.~Dai, H.~Li, J.~Song, X.~Wu, and A.~G. Hauptmann, ``Rethinking
  spatial invariance of convolutional networks for object counting,'' in
  \emph{Proceedings of the IEEE/CVF Conference on Computer Vision and Pattern
  Recognition}, 2022, pp. 19\,638--19\,648.

\bibitem{wang2022crowd}
Q.~Wang and T.~P. Breckon, ``Crowd counting via segmentation guided attention
  networks and curriculum loss,'' \emph{IEEE Transactions on Intelligent
  Transportation Systems}, 2022.

\bibitem{MSFANet}
P.~Thanasutives, K.-i. Fukui, M.~Numao, and B.~Kijsirikul, ``Encoder-decoder
  based convolutional neural networks with multi-scale-aware modules for crowd
  counting,'' in \emph{2020 25th International Conference on Pattern
  Recognition (ICPR)}.\hskip 1em plus 0.5em minus 0.4em\relax IEEE, 2021, pp.
  2382--2389.

\bibitem{gao2020learning}
J.~Gao, T.~Han, Y.~Yuan, and Q.~Wang, ``Learning independent instance maps for
  crowd localization,'' \emph{arXiv preprint arXiv:2012.04164}, 2020.

\bibitem{wang2021self}
Y.~Wang, J.~Hou, X.~Hou, and L.-P. Chau, ``A self-training approach for
  point-supervised object detection and counting in crowds,'' \emph{IEEE
  Transactions on Image Processing}, vol.~30, pp. 2876--2887, 2021.

\bibitem{song2021rethinking}
Q.~Song, C.~Wang, Z.~Jiang, Y.~Wang, Y.~Tai, C.~Wang, J.~Li, F.~Huang, and
  Y.~Wu, ``Rethinking counting and localization in crowds: A purely point-based
  framework,'' in \emph{Proceedings of the IEEE/CVF International Conference on
  Computer Vision}, 2021, pp. 3365--3374.

\bibitem{wang2021dense}
Y.~Wang, X.~Hou, and L.-P. Chau, ``Dense point prediction: A simple baseline
  for crowd counting and localization,'' in \emph{2021 IEEE International
  Conference on Multimedia \& Expo Workshops (ICMEW)}.\hskip 1em plus 0.5em
  minus 0.4em\relax IEEE, 2021, pp. 1--6.

\bibitem{Transformer_complex}
S.~Zhai, W.~Talbott, N.~Srivastava, C.~Huang, H.~Goh, R.~Zhang, and
  J.~Susskind, ``An attention free transformer,'' \emph{arXiv preprint
  arXiv:2105.14103}, 2021.

\bibitem{ASPP}
L.-C. Chen, G.~Papandreou, I.~Kokkinos, K.~Murphy, and A.~L. Yuille, ``Deeplab:
  Semantic image segmentation with deep convolutional nets, atrous convolution,
  and fully connected crfs,'' \emph{IEEE transactions on pattern analysis and
  machine intelligence}, vol.~40, no.~4, pp. 834--848, 2017.

\bibitem{cnn_position1}
M.~A. Islam, M.~Kowal, S.~Jia, K.~G. Derpanis, and N.~D. Bruce, ``Position,
  padding and predictions: A deeper look at position information in cnns,''
  \emph{arXiv preprint arXiv:2101.12322}, 2021.

\bibitem{cnn_position2}
M.~A. Islam, S.~Jia, and N.~D. Bruce, ``How much position information do
  convolutional neural networks encode?'' \emph{arXiv preprint
  arXiv:2001.08248}, 2020.

\bibitem{position_encodings}
X.~Chu, B.~Zhang, Z.~Tian, X.~Wei, and H.~Xia, ``Do we really need explicit
  position encodings for vision transformers?'' \emph{arXiv e-prints}, pp.
  arXiv--2102, 2021.

\bibitem{transcrowd}
D.~Liang, X.~Chen, W.~Xu, Y.~Zhou, and X.~Bai, ``Transcrowd: weakly-supervised
  crowd counting with transformers,'' \emph{Science China Information
  Sciences}, vol.~65, no.~6, pp. 1--14, 2022.

\bibitem{RANet}
A.~Zhang, J.~Shen, Z.~Xiao, F.~Zhu, X.~Zhen, X.~Cao, and L.~Shao, ``Relational
  attention network for crowd counting,'' in \emph{Proceedings of the IEEE/CVF
  International Conference on Computer Vision}, 2019, pp. 6788--6797.

\bibitem{MAN}
H.~Lin, Z.~Ma, R.~Ji, Y.~Wang, and X.~Hong, ``Boosting crowd counting via
  multifaceted attention,'' \emph{arXiv preprint arXiv:2203.02636}, 2022.

\bibitem{Transformer_Object_Tracking2}
N.~Wang, W.~Zhou, J.~Wang, and H.~Li, ``Transformer meets tracker: Exploiting
  temporal context for robust visual tracking,'' in \emph{2021 IEEE/CVF
  Conference on Computer Vision and Pattern Recognition (CVPR)}, 2021, pp.
  1571--1580.

\bibitem{Transformer_Object_Tracking1}
S.~Jiayao, S.~Zhou, Y.~Cui, and Z.~Fang, ``Real-time 3d single object tracking
  with transformer,'' \emph{IEEE Transactions on Multimedia}, pp. 1--1, 2022.

\bibitem{Swin_Transformer}
Z.~Liu, Y.~Lin, Y.~Cao, H.~Hu, Y.~Wei, Z.~Zhang, S.~Lin, and B.~Guo, ``Swin
  transformer: Hierarchical vision transformer using shifted windows,'' in
  \emph{2021 IEEE/CVF International Conference on Computer Vision (ICCV)},
  2021, pp. 9992--10\,002.

\bibitem{SETR}
S.~Zheng, J.~Lu, H.~Zhao, X.~Zhu, Z.~Luo, Y.~Wang, Y.~Fu, J.~Feng, T.~Xiang,
  P.~H. Torr, and L.~Zhang, ``Rethinking semantic segmentation from a
  sequence-to-sequence perspective with transformers,'' in \emph{2021 IEEE/CVF
  Conference on Computer Vision and Pattern Recognition (CVPR)}, 2021, pp.
  6877--6886.

\bibitem{ViT}
A.~Dosovitskiy, L.~Beyer, A.~Kolesnikov, D.~Weissenborn, X.~Zhai,
  T.~Unterthiner, M.~Dehghani, M.~Minderer, G.~Heigold, S.~Gelly \emph{et~al.},
  ``An image is worth 16x16 words: Transformers for image recognition at
  scale,'' \emph{arXiv preprint arXiv:2010.11929}, 2020.

\bibitem{DETR}
N.~Carion, F.~Massa, G.~Synnaeve, N.~Usunier, A.~Kirillov, and S.~Zagoruyko,
  ``End-to-end object detection with transformers,'' in \emph{European
  conference on computer vision}.\hskip 1em plus 0.5em minus 0.4em\relax
  Springer, 2020, pp. 213--229.

\bibitem{NIPS2017_3f5ee243}
\BIBentryALTinterwordspacing
A.~Vaswani, N.~Shazeer, N.~Parmar, J.~Uszkoreit, L.~Jones, A.~N. Gomez, L.~u.
  Kaiser, and I.~Polosukhin, ``Attention is all you need,'' in \emph{Advances
  in Neural Information Processing Systems}, I.~Guyon, U.~V. Luxburg,
  S.~Bengio, H.~Wallach, R.~Fergus, S.~Vishwanathan, and R.~Garnett, Eds.,
  vol.~30.\hskip 1em plus 0.5em minus 0.4em\relax Curran Associates, Inc.,
  2017. [Online]. Available:
  \url{https://proceedings.neurips.cc/paper/2017/file/3f5ee243547dee91fbd053c1c4a845aa-Paper.pdf}
\BIBentrySTDinterwordspacing

\bibitem{shang2016end}
C.~Shang, H.~Ai, and B.~Bai, ``End-to-end crowd counting via joint learning
  local and global count,'' in \emph{2016 IEEE International Conference on
  Image Processing (ICIP)}.\hskip 1em plus 0.5em minus 0.4em\relax IEEE, 2016,
  pp. 1215--1219.

\bibitem{Cnn-based}
V.~A. Sindagi and V.~M. Patel, ``Cnn-based cascaded multi-task learning of
  high-level prior and density estimation for crowd counting,'' in \emph{2017
  14th IEEE international conference on advanced video and signal based
  surveillance (AVSS)}.\hskip 1em plus 0.5em minus 0.4em\relax IEEE, 2017, pp.
  1--6.

\bibitem{CrowdNet}
L.~Boominathan, S.~S. Kruthiventi, and R.~V. Babu, ``Crowdnet: A deep
  convolutional network for dense crowd counting,'' in \emph{Proceedings of the
  24th ACM international conference on Multimedia}, 2016, pp. 640--644.

\bibitem{DADNet}
D.~Guo, K.~Li, Z.-J. Zha, and M.~Wang, ``Dadnet: Dilated-attention-deformable
  convnet for crowd counting,'' 10 2019, pp. 1823--1832.

\bibitem{DecideNet}
J.~Liu, C.~Gao, D.~Meng, and A.~G. Hauptmann, ``Decidenet: Counting varying
  density crowds through attention guided detection and density estimation,''
  in \emph{2018 IEEE/CVF Conference on Computer Vision and Pattern
  Recognition}, 2018, pp. 5197--5206.

\bibitem{NoiseCC}
J.~Wan and A.~Chan, ``Modeling noisy annotations for crowd counting,''
  \emph{Advances in Neural Information Processing Systems}, vol.~33, 2020.

\bibitem{Alpher53}
S.~Ren, K.~He, R.~Girshick, and J.~Sun, ``Faster r-cnn: Towards real-time
  object detection with region proposal networks,'' \emph{IEEE Transactions on
  Pattern Analysis and Machine Intelligence}, vol.~39, no.~6, pp. 1137--1149,
  2017.

\bibitem{Alpher54}
J.~Redmon, S.~Divvala, R.~Girshick, and A.~Farhadi, ``You only look once:
  Unified, real-time object detection,'' in \emph{2016 IEEE Conference on
  Computer Vision and Pattern Recognition (CVPR)}, 2016, pp. 779--788.

\bibitem{Alpher55}
\BIBentryALTinterwordspacing
T.~N. Mundhenk, G.~Konjevod, W.~A. Sakla, and K.~Boakye, ``A large contextual
  dataset for classification, detection and counting of cars with deep
  learning,'' \emph{CoRR}, vol. abs/1609.04453, 2016. [Online]. Available:
  \url{http://arxiv.org/abs/1609.04453}
\BIBentrySTDinterwordspacing

\bibitem{Alpher56}
J.~Redmon and A.~Farhadi, ``Yolo9000: Better, faster, stronger,'' in \emph{2017
  IEEE Conference on Computer Vision and Pattern Recognition (CVPR)}, 2017, pp.
  6517--6525.

\bibitem{Alpher57}
T.-Y. Lin, P.~Goyal, R.~Girshick, K.~He, and P.~Dollár, ``Focal loss for dense
  object detection,'' in \emph{2017 IEEE International Conference on Computer
  Vision (ICCV)}, 2017, pp. 2999--3007.

\bibitem{Alpher58}
------, ``Focal loss for dense object detection,'' \emph{IEEE Transactions on
  Pattern Analysis and Machine Intelligence}, vol.~42, no.~2, pp. 318--327,
  2020.

\bibitem{Alpher59}
T.~Stahl, S.~L. Pintea, and J.~C. van Gemert, ``Divide and count: Generic
  object counting by image divisions,'' \emph{IEEE Transactions on Image
  Processing}, vol.~28, no.~2, pp. 1035--1044, 2019.

\bibitem{Alpher60}
E.~Goldman, R.~Herzig, A.~Eisenschtat, J.~Goldberger, and T.~Hassner, ``Precise
  detection in densely packed scenes,'' in \emph{2019 IEEE/CVF Conference on
  Computer Vision and Pattern Recognition (CVPR)}, 2019, pp. 5222--5231.

\bibitem{Alpher50}
X.~Liu, J.~van~de Weijer, and A.~D. Bagdanov, ``Leveraging unlabeled data for
  crowd counting by learning to rank,'' in \emph{2018 IEEE/CVF Conference on
  Computer Vision and Pattern Recognition}, 2018, pp. 7661--7669.

\bibitem{Alpher52}
I.~H. Laradji, N.~Rostamzadeh, P.~O. Pinheiro, D.~Vazquez, and M.~Schmidt,
  ``Where are the blobs: Counting by localization with point supervision,''
  2018.

\bibitem{PSDDN}
Y.~Liu, M.~Shi, Q.~Zhao, and X.~Wang, ``Point in, box out: Beyond counting
  persons in crowds,'' in \emph{2019 IEEE/CVF Conference on Computer Vision and
  Pattern Recognition (CVPR)}, 2019, pp. 6462--6471.

\bibitem{DSSINet}
V.~K. Valloli and K.~Mehta, ``W-net: Reinforced u-net for density map
  estimation,'' 2019.

\bibitem{DOTA}
J.~Ding, N.~Xue, G.-S. Xia, X.~Bai, W.~Yang, M.~Y. Yang, S.~Belongie, J.~Luo,
  M.~Datcu, M.~Pelillo, and L.~Zhang, ``Object detection in aerial images: A
  large-scale benchmark and challenges,'' 2021.

\bibitem{CARPK}
M.-R. Hsieh, Y.-L. Lin, and W.~H. Hsu, ``Drone-based object counting by
  spatially regularized regional proposal network,'' in \emph{2017 IEEE
  International Conference on Computer Vision (ICCV)}, 2017, pp. 4165--4173.

\bibitem{NWPU}
Q.~Wang, J.~Gao, W.~Lin, and X.~Li, ``Nwpu-crowd: A large-scale benchmark for
  crowd counting and localization,'' \emph{IEEE transactions on pattern
  analysis and machine intelligence}, vol.~43, no.~6, pp. 2141--2149, 2020.

\bibitem{correntropy2}
R.~He, W.-S. Zheng, and B.-G. Hu, ``Maximum correntropy criterion for robust
  face recognition,'' \emph{IEEE Transactions on Pattern Analysis and Machine
  Intelligence}, vol.~33, no.~8, pp. 1561--1576, 2010.

\bibitem{correntropy3}
W.~Liu, P.~P. Pokharel, and J.~C. Principe, ``Correntropy: Properties and
  applications in non-gaussian signal processing,'' \emph{IEEE Transactions on
  signal processing}, vol.~55, no.~11, pp. 5286--5298, 2007.

\bibitem{V3}
R.~Guerrero-G\'omez-Olmedo, B.~Torre-Jim\'enez, R.~L\'opez-Sastre,
  S.~Maldonado-Basc\'on, and D.~O. noro Rubio, ``Extremely overlapping vehicle
  counting,'' in \emph{Iberian Conference on Pattern Recognition and Image
  Analysis (IbPRIA)}, 2015.

\bibitem{TCSVT1}
Z.~Yan, P.~Li, B.~Wang, D.~Ren, and W.~Zuo, ``Towards learning multi-domain
  crowd counting,'' \emph{IEEE Transactions on Circuits and Systems for Video
  Technology}, pp. 1--1, 2021.

\bibitem{TCSVT2}
U.~Sajid, H.~Sajid, H.~Wang, and G.~Wang, ``Zoomcount: A zooming mechanism for
  crowd counting in static images,'' \emph{IEEE Transactions on Circuits and
  Systems for Video Technology}, vol.~30, no.~10, pp. 3499--3512, 2020.

\bibitem{TCSVT3}
J.~Gao, Q.~Wang, and X.~Li, ``Pcc net: Perspective crowd counting via spatial
  convolutional network,'' \emph{IEEE Transactions on Circuits and Systems for
  Video Technology}, vol.~30, no.~10, pp. 3486--3498, 2020.

\bibitem{TCSVT4}
M.~Zhao, C.~Zhang, J.~Zhang, F.~Porikli, B.~Ni, and W.~Zhang, ``Scale-aware
  crowd counting via depth-embedded convolutional neural networks,'' \emph{IEEE
  Transactions on Circuits and Systems for Video Technology}, vol.~30, no.~10,
  pp. 3651--3662, 2020.

\bibitem{TCSVT5}
H.~Zheng, Z.~Lin, J.~Cen, Z.~Wu, and Y.~Zhao, ``Cross-line pedestrian counting
  based on spatially-consistent two-stage local crowd density estimation and
  accumulation,'' \emph{IEEE Transactions on Circuits and Systems for Video
  Technology}, vol.~29, no.~3, pp. 787--799, 2019.

\bibitem{TCSVT6}
S.~Jiang, X.~Lu, Y.~Lei, and L.~Liu, ``Mask-aware networks for crowd
  counting,'' \emph{IEEE Transactions on Circuits and Systems for Video
  Technology}, vol.~30, no.~9, pp. 3119--3129, 2020.

\bibitem{TCSVT7}
B.~Sheng, C.~Shen, G.~Lin, J.~Li, W.~Yang, and C.~Sun, ``Crowd counting via
  weighted vlad on a dense attribute feature map,'' \emph{IEEE Transactions on
  Circuits and Systems for Video Technology}, vol.~28, no.~8, pp. 1788--1797,
  2018.

\bibitem{TNNLS1}
X.~Jiang, L.~Zhang, P.~Lv, Y.~Guo, R.~Zhu, Y.~Li, Y.~Pang, X.~Li, B.~Zhou, and
  M.~Xu, ``Learning multi-level density maps for crowd counting,'' \emph{IEEE
  Transactions on Neural Networks and Learning Systems}, vol.~31, no.~8, pp.
  2705--2715, 2020.

\bibitem{TNNLS2}
J.~Gao, T.~Han, Y.~Yuan, and Q.~Wang, ``Domain-adaptive crowd counting via
  high-quality image translation and density reconstruction,'' \emph{IEEE
  Transactions on Neural Networks and Learning Systems}, pp. 1--13, 2021.

\bibitem{TNNLS3}
Q.~Wang, T.~Han, J.~Gao, and Y.~Yuan, ``Neuron linear transformation: Modeling
  the domain shift for crowd counting,'' \emph{IEEE Transactions on Neural
  Networks and Learning Systems}, pp. 1--13, 2021.

\bibitem{TNNLS4}
Y.~Luo, J.~Lü, X.~Jiang, and B.~Zhang, ``Learning from architectural
  redundancy: Enhanced deep supervision in deep multipath encoder-decoder
  networks,'' \emph{IEEE Transactions on Neural Networks and Learning Systems},
  pp. 1--14, 2021.

\bibitem{TNNLS5}
X.~Chen, X.~Chen, Y.~Zhang, X.~Fu, and Z.-J. Zha, ``Laplacian pyramid neural
  network for dense continuous-value regression for complex scenes,''
  \emph{IEEE Transactions on Neural Networks and Learning Systems}, vol.~32,
  no.~11, pp. 5034--5046, 2021.

\bibitem{BL}
Z.~Ma, X.~Wei, X.~Hong, and Y.~Gong, ``Bayesian loss for crowd count estimation
  with point supervision,'' in \emph{Proceedings of the IEEE/CVF International
  Conference on Computer Vision (ICCV)}, October 2019.

\bibitem{MCNN}
Y.~Zhang, D.~Zhou, S.~Chen, S.~Gao, and Y.~Ma, ``Single-image crowd counting
  via multi-column convolutional neural network,'' in \emph{Proceedings of the
  IEEE Conference on Computer Vision and Pattern Recognition (CVPR)}, June
  2016.

\bibitem{cross-scenes}
C.~Zhang, H.~Li, X.~Wang, and X.~Yang, ``Cross-scene crowd counting via deep
  convolutional neural networks,'' in \emph{Proceedings of the IEEE Conference
  on Computer Vision and Pattern Recognition (CVPR)}, June 2015.

\bibitem{Adaptive-Density-Map1}
J.~Wan and A.~Chan, ``Adaptive density map generation for crowd counting,'' in
  \emph{Proceedings of the IEEE/CVF International Conference on Computer Vision
  (ICCV)}, October 2019.

\bibitem{Adaptive-Density-Map2}
J.~Wan, Q.~Wang, and A.~B. Chan, ``Kernel-based density map generation for
  dense object counting,'' \emph{IEEE Transactions on Pattern Analysis and
  Machine Intelligence}, 2020.

\bibitem{DUT-OMRON}
C.~Yang and Zhang, ``Saliency detection via graph-based manifold ranking,'' in
  \emph{Computer Vision and Pattern Recognition (CVPR), 2013 IEEE Conference
  on}.\hskip 1em plus 0.5em minus 0.4em\relax IEEE, 2013, pp. 3166--3173.

\bibitem{attention-psychology1}
L.~Itti, C.~Koch, and E.~Niebur, ``A model of saliency-based visual attention
  for rapid scene analysis,'' \emph{IEEE Transactions on Pattern Analysis and
  Machine Intelligence}, vol.~20, no.~11, pp. 1254--1259, 1998.

\bibitem{attention-psychology2}
\BIBentryALTinterwordspacing
A.~M. Treisman and G.~Gelade, ``A feature-integration theory of attention,''
  \emph{Cognitive Psychology}, vol.~12, no.~1, pp. 97--136, 1980. [Online].
  Available:
  \url{https://www.sciencedirect.com/science/article/pii/0010028580900055}
\BIBentrySTDinterwordspacing

\bibitem{attention-psychology3}
C.~Koch and S.~Ullman, \emph{Shifts in Selective Visual Attention: Towards the
  Underlying Neural Circuitry}.\hskip 1em plus 0.5em minus 0.4em\relax
  Dordrecht: Springer Netherlands, 1987, pp. 115--141.

\bibitem{Salient-Object-Detection}
W.~Wang, Q.~Lai, H.~Fu, J.~Shen, H.~Ling, and R.~Yang, ``Salient object
  detection in the deep learning era: An in-depth survey,'' \emph{IEEE
  Transactions on Pattern Analysis and Machine Intelligence}, pp. 1--1, 2021.

\bibitem{JHU1}
V.~A. Sindagi, R.~Yasarla, and V.~M. Patel, ``Pushing the frontiers of
  unconstrained crowd counting: New dataset and benchmark method,'' in
  \emph{Proceedings of the IEEE International Conference on Computer Vision},
  2019, pp. 1221--1231.

\bibitem{JHU2}
------, ``Jhu-crowd++: Large-scale crowd counting dataset and a benchmark
  method,'' \emph{Technical Report}, 2020.

\bibitem{traditional-detetcion1}
M.~Li, Z.~Zhang, K.~Huang, and T.~Tan, ``Estimating the number of people in
  crowded scenes by mid based foreground segmentation and head-shoulder
  detection,'' in \emph{2008 19th International Conference on Pattern
  Recognition}, 2008, pp. 1--4.

\bibitem{traditional-detetcion2}
W.~Ge and R.~T. Collins, ``Marked point processes for crowd counting,'' in
  \emph{2009 IEEE Conference on Computer Vision and Pattern Recognition}, 2009,
  pp. 2913--2920.

\bibitem{traditional-regression1}
A.~B. Chan, Z.-S.~J. Liang, and N.~Vasconcelos, ``Privacy preserving crowd
  monitoring: Counting people without people models or tracking,'' in
  \emph{2008 IEEE Conference on Computer Vision and Pattern Recognition}, 2008,
  pp. 1--7.

\bibitem{traditional-regression2}
A.~B. Chan and N.~Vasconcelos, ``Bayesian poisson regression for crowd
  counting,'' in \emph{2009 IEEE 12th International Conference on Computer
  Vision}, 2009, pp. 545--551.

\bibitem{traditional-regression3}
H.~Idrees, I.~Saleemi, C.~Seibert, and M.~Shah, ``Multi-source multi-scale
  counting in extremely dense crowd images,'' in \emph{2013 IEEE Conference on
  Computer Vision and Pattern Recognition}, 2013, pp. 2547--2554.

\bibitem{first-denisty}
V.~Lempitsky and A.~Zisserman, ``Learning to count objects in images.'' 01
  2010, pp. 1324--1332.

\bibitem{SASNet}
Q.~Song, C.~Wang, Y.~Wang, Y.~Tai, C.~Wang, J.~Li, J.~Wu, and J.~Ma, ``To
  choose or to fuse? scale selection for crowd counting,'' in \emph{Proceedings
  of the AAAI Conference on Artificial Intelligence}, vol.~35, no.~3, 2021, pp.
  2576--2583.

\bibitem{switch-CNN}
D.~Babu~Sam, S.~Surya, and R.~Venkatesh~Babu, ``Switching convolutional neural
  network for crowd counting,'' in \emph{Proceedings of the IEEE Conference on
  Computer Vision and Pattern Recognition (CVPR)}, July 2017.

\bibitem{scale1}
X.~Jiang, Z.~Xiao, B.~Zhang, X.~Zhen, X.~Cao, D.~Doermann, and L.~Shao, ``Crowd
  counting and density estimation by trellis encoder-decoder networks,'' in
  \emph{Proceedings of the IEEE/CVF Conference on Computer Vision and Pattern
  Recognition (CVPR)}, June 2019.

\bibitem{fuse}
V.~A. Sindagi and V.~M. Patel, ``Multi-level bottom-top and top-bottom feature
  fusion for crowd counting,'' in \emph{Proceedings of the IEEE/CVF
  International Conference on Computer Vision (ICCV)}, October 2019.

\bibitem{DM-general}
J.~Wan, Z.~Liu, and A.~B. Chan, ``A generalized loss function for crowd
  counting and localization,'' in \emph{Proceedings of the IEEE/CVF Conference
  on Computer Vision and Pattern Recognition (CVPR)}, June 2021, pp.
  1974--1983.

\bibitem{ADSCNet}
S.~Bai, Z.~He, Y.~Qiao, H.~Hu, W.~Wu, and J.~Yan, ``Adaptive dilated network
  with self-correction supervision for counting,'' in \emph{Proceedings of the
  IEEE/CVF Conference on Computer Vision and Pattern Recognition (CVPR)}, June
  2020.

\bibitem{OT-original}
C.~Villani, \emph{Optimal transport: old and new}.\hskip 1em plus 0.5em minus
  0.4em\relax Springer, 2009, vol. 338.

\bibitem{OT-theory}
\BIBentryALTinterwordspacing
G.~Peyré and M.~Cuturi, ``Computational optimal transport: With applications
  to data science,'' \emph{Foundations and Trends® in Machine Learning},
  vol.~11, no. 5-6, pp. 355--607, 2019. [Online]. Available:
  \url{http://dx.doi.org/10.1561/2200000073}
\BIBentrySTDinterwordspacing

\bibitem{Correntropy-1}
B.~Chen, Y.~Xie, X.~Wang, Z.~Yuan, P.~Ren, and J.~Qin, ``Multikernel
  correntropy for robust learning,'' \emph{IEEE Transactions on Cybernetics},
  pp. 1--12, 2021.

\bibitem{Pix2Pix-GAN}
P.~Isola, J.-Y. Zhu, T.~Zhou, and A.~A. Efros, ``Image-to-image translation
  with conditional adversarial networks,'' in \emph{Proceedings of the IEEE
  Conference on Computer Vision and Pattern Recognition (CVPR)}, July 2017.

\bibitem{UCF-CC-50}
A.~Bansal and K.~S. Venkatesh, ``People counting in high density crowds from
  still images,'' 2015.

\bibitem{UCF-QNRF}
H.~Idrees, M.~Tayyab, K.~Athrey, D.~Zhang, S.~Al-Maadeed, N.~Rajpoot, and
  M.~Shah, ``Composition loss for counting, density map estimation and
  localization in dense crowds,'' in \emph{Proceedings of the European
  Conference on Computer Vision (ECCV)}, September 2018.

\bibitem{SANet}
X.~Cao, Z.~Wang, Y.~Zhao, and F.~Su, ``Scale aggregation network for accurate
  and efficient crowd counting,'' in \emph{Proceedings of the European
  Conference on Computer Vision (ECCV)}, September 2018.

\bibitem{CSRNet}
Y.~Li, X.~Zhang, and D.~Chen, ``Csrnet: Dilated convolutional neural networks
  for understanding the highly congested scenes,'' in \emph{Proceedings of the
  IEEE conference on computer vision and pattern recognition}, 2018, pp.
  1091--1100.

\bibitem{CAN}
W.~Liu, M.~Salzmann, and P.~Fua, ``Context-aware crowd counting,'' in
  \emph{Proceedings of the IEEE/CVF Conference on Computer Vision and Pattern
  Recognition}, 2019, pp. 5099--5108.

\bibitem{MBTTBF}
V.~A. Sindagi and V.~M. Patel, ``Multi-level bottom-top and top-bottom feature
  fusion for crowd counting,'' in \emph{Proceedings of the IEEE/CVF
  International Conference on Computer Vision}, 2019, pp. 1002--1012.

\bibitem{ADCrowdNet}
N.~Liu, Y.~Long, C.~Zou, Q.~Niu, L.~Pan, and H.~Wu, ``Adcrowdnet: An
  attention-injective deformable convolutional network for crowd
  understanding,'' in \emph{Proceedings of the IEEE/CVF Conference on Computer
  Vision and Pattern Recognition}, 2019, pp. 3225--3234.

\bibitem{LSC-CNN}
D.~B. Sam, S.~V. Peri, M.~N. Sundararaman, A.~Kamath, and V.~B. Radhakrishnan,
  ``Locate, size and count: Accurately resolving people in dense crowds via
  detection,'' \emph{IEEE transactions on pattern analysis and machine
  intelligence}, 2020.

\bibitem{SFCN}
Q.~Wang, J.~Gao, W.~Lin, and Y.~Yuan, ``Learning from synthetic data for crowd
  counting in the wild,'' in \emph{Proceedings of the IEEE/CVF Conference on
  Computer Vision and Pattern Recognition}, 2019, pp. 8198--8207.

\bibitem{DM-Count}
B.~Wang, H.~Liu, D.~Samaras, and M.~Hoai, ``Distribution matching for crowd
  counting,'' \emph{arXiv preprint arXiv:2009.13077}, 2020.

\bibitem{AMRNet}
X.~Liu, J.~Yang, W.~Ding, T.~Wang, Z.~Wang, and J.~Xiong, ``Adaptive mixture
  regression network with local counting map for crowd counting,'' in
  \emph{Computer Vision--ECCV 2020: 16th European Conference, Glasgow, UK,
  August 23--28, 2020, Proceedings, Part XXIV 16}.\hskip 1em plus 0.5em minus
  0.4em\relax Springer, 2020, pp. 241--257.

\bibitem{LibraNet}
L.~Liu, H.~Lu, H.~Zou, H.~Xiong, Z.~Cao, and C.~Shen, ``Weighing counts:
  Sequential crowd counting by reinforcement learning,'' in \emph{European
  Conference on Computer Vision}.\hskip 1em plus 0.5em minus 0.4em\relax
  Springer, 2020, pp. 164--181.

\bibitem{CMTL}
V.~A. Sindagi and V.~M. Patel, ``Cnn-based cascaded multi-task learning of
  high-level prior and density estimation for crowd counting,'' in \emph{2017
  14th IEEE International Conference on Advanced Video and Signal Based
  Surveillance (AVSS)}.\hskip 1em plus 0.5em minus 0.4em\relax IEEE, 2017, pp.
  1--6.

\bibitem{ADMG}
J.~Wan, Q.~Wang, and A.~B. Chan, ``Kernel-based density map generation for
  dense object counting,'' \emph{IEEE Transactions on Pattern Analysis and
  Machine Intelligence}, pp. 1--1, 2020.

\bibitem{SaCNN}
L.~Zhang, M.~Shi, and Q.~Chen, ``Crowd counting via scale-adaptive
  convolutional neural network,'' in \emph{2018 IEEE Winter Conference on
  Applications of Computer Vision (WACV)}.\hskip 1em plus 0.5em minus
  0.4em\relax IEEE, 2018, pp. 1113--1121.

\bibitem{GL}
J.~Wan, Z.~Liu, and A.~B. Chan, ``A generalized loss function for crowd
  counting and localization,'' in \emph{2021 IEEE/CVF Conference on Computer
  Vision and Pattern Recognition (CVPR)}, 2021, pp. 1974--1983.

\bibitem{UOT}
Z.~Ma, X.~Wei, X.~Hong, H.~Lin, Y.~Qiu, and Y.~Gong, ``Learning to count via
  unbalanced optimal transport,'' in \emph{Proceedings of the AAAI Conference
  on Artificial Intelligence}, vol.~35, no.~3, 2021, pp. 2319--2327.

\bibitem{KMPE1}
Y.~Yang, D.~Fan, S.~Du, M.~Wang, B.~Chen, and Y.~Gao, ``Point set registration
  with similarity and affine transformations based on bidirectional kmpe
  loss,'' \emph{IEEE transactions on cybernetics}, vol.~51, no.~3, pp.
  1678--1689, 2019.

\bibitem{KMPE2}
B.~Chen, L.~Xing, X.~Wang, J.~Qin, and N.~Zheng, ``Robust learning with kernel
  mean $ p $-power error loss,'' \emph{IEEE transactions on cybernetics},
  vol.~48, no.~7, pp. 2101--2113, 2017.

\bibitem{M-SFANet}
P.~Thanasutives, K.-i. Fukui, M.~Numao, and B.~Kijsirikul, ``Encoder-decoder
  based convolutional neural networks with multi-scale-aware modules for crowd
  counting,'' in \emph{2020 25th International Conference on Pattern
  Recognition (ICPR)}.\hskip 1em plus 0.5em minus 0.4em\relax IEEE, 2021, pp.
  2382--2389.

\end{thebibliography}
\end{document}